\documentclass{article}

\PassOptionsToPackage{numbers, compress}{natbib}

\usepackage[preprint]{neurips_2024}

\usepackage[utf8]{inputenc} 
\usepackage[T1]{fontenc}    
\usepackage{hyperref}       
\usepackage{url}            
\usepackage{booktabs}       
\usepackage{amsfonts}       
\usepackage{nicefrac}       
\usepackage{microtype}      
\usepackage{xcolor}         
\usepackage{wrapfig}
\usepackage{amssymb} 
\usepackage[nointegrals]{wasysym} 
\usepackage{tikz}
\usepackage{amsfonts}
\usepackage{bm}
\usepackage{color}
\usepackage{enumitem}
\usepackage{float}
\usepackage{graphicx}
\usepackage{hyperref}
\usepackage{mathtools,soul}
\usepackage{multicol}
\usepackage{multirow}
\usepackage{soul}
\usepackage{caption}
\usepackage{subcaption}
\usepackage{xspace}
\usepackage{pifont}
\usepackage{siunitx}

\usepackage{algorithm}
\usepackage{algorithmic}
\newtheorem{theorem}{Theorem}[section]

\newtheorem{definition}[theorem]{Definition}

\newcommand{\RNum}[1]{\uppercase\expandafter{\romannumeral #1\relax}}
\newcolumntype{d}[1]{S[table-format=#1]}

\definecolor{myblue}{rgb}{0.651, 0.808, 0.890}

\title{Towards Subgraph Isomorphism Counting with \\
Graph Kernels}

\author{Xin Liu, Weiqi Wang, Jiaxin Bai, Yangqiu Song \\
Department of Computer Science and Engineering \\
Hong Kong University of Science and Technology \\
Hong Kong, China \\
\texttt{\{xliucr,wwangbw,jbai,yqsong\}@cse.ust.hk}
}

\begin{document}

\maketitle

\begin{abstract}
Subgraph isomorphism counting is known as \#P-complete and requires exponential time to find the accurate solution. Utilizing representation learning has been shown as a promising direction to represent substructures and approximate the solution. Graph kernels that implicitly capture the correlations among substructures in diverse graphs have exhibited great discriminative power in graph classification, so we pioneeringly investigate their potential in counting subgraph isomorphisms and further explore the augmentation of kernel capability through various variants, including polynomial and Gaussian kernels. Through comprehensive analysis, we enhance the graph kernels by incorporating neighborhood information. Finally, we present the results of extensive experiments to demonstrate the effectiveness of the enhanced graph kernels and discuss promising directions for future research.
\end{abstract}

\section{Introduction}
\label{sec:intro}

The objective of subgraph isomorphism counting is to determine the number of subgraphs in a given graph that match a specific \textit{pattern} graph, i.e., that are isomorphic to it.
This technique is highly valuable in knowledge discovery and data mining applications, such as identifying protein interactions in bioinformatics~\citep{milo2002network,alon2008biomolecular}.
Moreover, it is beneficial for analyzing heterogeneous information networks (HINs), including knowledge graphs~\citep{shen2019multi}, online social networks~\citep{kuramochi2004grew}, and recommendation systems~\citep{huang2016meta, zhao2017meta}.
The diverse range of types within the HIN schema offers meaningful semantics, making subgraph counting a valuable component in various types of queries.

Numerous backtracking algorithms and indexing-based approaches have been proposed to tackle the challenges of subgraph isomorphisms~\citep{ullmann1976an, vento2004a, he2008graphs, han2013turbo, carletti2018challenging, klein2011ctindex}.
However, previous research on similar tasks often focuses on specific constraints, with limited discussions on general patterns in heterogeneous graphs.
Due to the NP-hard nature of subgraph isomorphisms, researchers have also explored efficient approximations of the number of subgraph isomorphisms instead of exact counts, using techniques such as sampling~\citep{, jha2015path} and color coding~\citep{zhao2012sahad}.
While these approximate solutions are relatively efficient, generalizing them to heterogeneous settings is challenging, especially considering the high memory consumption and dynamic programming complexity in heuristic rules.

Graph-based learning has recently gained significant interest, and graph kernel methods have been extensively applied in machine learning for graph classification~\citep{kashima2002kernels, glavavs2013recognizing, zhang2013fast, jie2014topological} and clustering~\citep{clariso2018applying, tepeli2020pamogk}.
These applications involve more \textit{local} decisions, where learning algorithms typically make inferences by examining the local structures of a graph.
Certain kernels are designed to capture the structural information of graphs, such as the Weisfeiler-Leman subtree kernel (WL kernel)~\citep{shervashidze2011weisfeiler}, which naturally lends itself to isomorphism testing~\citep{weisfeiler1968reduction}.
However, there exists a gap between isomorphism testing and subgraph isomorphism counting: the former is a binary problem, while the latter is a \#P problem.
Furthermore, isomorphism testing only considers the global structure histograms, whereas subgraph isomorphism counting requires analyzing the local structure combinations.
Nonetheless, we can still utilize graph kernels to approximate the number of isomorphic subgraphs using kernel values among thousands of graphs to represent substructures implicitly. This solution could be feasible because kernel functions and Gram matrix construction are cost-effective. With neighborhood-aware techniques and kernel tricks, we can further elevate the performance of graph kernels, making them comparable to neural networks.
\section{Related Work}
\label{sec:related}

Subgraph isomorphism search, which involves finding all identical bijections, poses a more challenging problem and has been proven to be NP-complete.
Numerous subgraph isomorphism algorithms have been developed, including backtracking-based algorithms and indexing-based algorithms.
Ullmann's algorithm~\citep{ullmann1976an} is the first and the most straightforward, which enumerates all possible mappings and prunes non-matching mappings as early as possible.
Several heuristic strategies have been proposed to reduce the search space and improve efficiency, such as VF2~\citep{vento2004a},
VF3~\citep{carletti2018challenging},
TurboIso~\citep{han2013turbo}, and BoostIso~\citep{ren2015exploiting}.
Some algorithms are specifically designed and optimized for particular applications and database engines
Graph query languages, such as GraphGrep~\citep{giugno2002graphgrep} and GraphQL~\citep{he2008graphs}, represent patterns as hash-based fingerprints and use overlapping label-paths in the depth-first-search tree to represent branches.
Various composition and selection operators can be designed based on graph structures and graph algebras.
Indexing techniques play a crucial role in this area, like gIndex~\citep{yan2004graph}, FG-Index~\citep{cheng2007fgindex}, and CT-Index~\citep{klein2011ctindex}.
Another significant direction is approximating the number of subgraph isomorphisms, striking a balance between accuracy and efficiency.
Sampling techniques~\citep{wernicke2005a, ribeiro2010efficient, jha2015path} and color coding~\citep{alon1995color, bressan2019motivo} are commonly employed.
However, most methods focus on homogeneous graphs and induced subgraphs, which limits their applications in real scenarios.

Graph neural networks (GNNs) can capture rich structural information of graphs, and researchers have explored their potential in subgraph matching.
The message passing framework is one such technique that leverages the representation of a neighborhood as a multiset of features and aggregates neighbor messages to find functional groups in chemical molecules~\citep{gilmer2017neural}.
Additionally, certain substructures in social networks enhance the effectiveness of recommender systems~\citep{ying2018graph}.
Subsequently, researchers have employed graph neural networks for subgraph counting and matching purposes.
\citet{liu2020neural} developed a comprehensive and unified end-to-end framework that combines sequence and graph models to count subgraph isomorphisms.
\citet{ying2020neural} utilized graph neural networks to embed vertices and employed a voting algorithm to match subgraphs using the acquired representations.
\citet{chen2020can} conducted a theoretical analysis of the upper limit of $k$-WL and similar message passing variants.

Given that the message-passing framework simulates the process of the Weisfeiler-Leman algorithm, it is still under research whether general graph kernels can be used to predict the numbers of subgraph isomorphism.
One of the mainstream paradigms in the design of graph kernels is to present and compare local structures.
The principal idea is to encode graphs into sparse vectors, and similar topologies should have similar representations. For example, the represented objects can be bags of components, e.g., triangles~\citep{shervashidze2009efficient}, paths~\citep{borgward2005shortest}, or neighborhood~\citep{shervashidze2011weisfeiler}.
However, simple structures have limited the discriminative power of classifiers because two different structures may result in the same representations.
Therefore, people explore higher-order structures~\citep{shervashidze2011weisfeiler, morris2020weisfeiler}.
Higher-order structures usually come with exponential costs, so many other graph kernels turn to generalization~\citep{schulz2022a} and efficiency~\citep{bause2022lgradual} with the loose of guidance.
\section{Background and Motivations}

\subsection{Problem Definition}
\label{sec:iso}

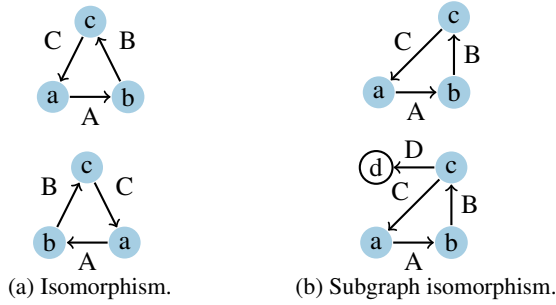
\begin{wrapfigure}{!r}{0.64\textwidth}
\vspace{-0.25in}
\begin{subfigure}{.40\linewidth}
        \centering
        \begin{tikzpicture}[thick,scale=0.5, every node/.style={scale=1.0}]
          \node[circle, fill=myblue, inner sep=2pt, minimum size=2pt] (a) at (0,0) {a};
          \node[circle, fill=myblue, inner sep=1.4pt, minimum size=2pt] (b) at (2,0) {b};
          \node[circle, fill=myblue, inner sep=2pt, minimum size=2pt] (c) at (1,2) {c};
        
          \draw[->] (a) -- node[below] {A} (b);
          \draw[->] (b) -- node[above right] {B} (c);
          \draw[->] (c) -- node[above left] {C} (a);
        \end{tikzpicture}
    
        \begin{tikzpicture}[thick,scale=0.5, every node/.style={scale=1.0}]
          \node[circle, fill=myblue, inner sep=2pt, minimum size=2pt] (a) at (2,0) {a};
          \node[circle, fill=myblue, inner sep=1.4pt, minimum size=2pt] (b) at (0,0) {b};
          \node[circle, fill=myblue, inner sep=2pt, minimum size=2pt] (c) at (1,2) {c};
          \node (d) at (1,2.6) {};
        
          \draw[->] (a) -- node[below] {A} (b);
          \draw[->] (b) -- node[above left] {B} (c);
          \draw[->] (c) -- node[above right] {C} (a);
        \end{tikzpicture}
        \vspace{-0.1in}
        \caption{Isomorphism.}
        \label{fig:preliminary_iso}
    \end{subfigure}
    \begin{subfigure}{.58\linewidth}
        \centering
        \begin{tikzpicture}[thick,scale=0.5, every node/.style={scale=1.0}]
          \node[circle, fill=myblue, inner sep=2pt, minimum size=2pt] (a) at (0,0) {a};
          \node[circle, fill=myblue, inner sep=1.4pt, minimum size=2pt] (b) at (2,0) {b};
          \node[circle, fill=myblue, inner sep=2pt, minimum size=2pt] (c) at (2,2) {c};
        
          \draw[->] (a) -- node[below] {A} (b);
          \draw[->] (b) -- node[right] {B} (c);
          \draw[->] (c) -- node[above left=-2pt] {C} (a);
        \end{tikzpicture}
    
        \begin{tikzpicture}[thick,scale=0.5, every node/.style={scale=1.0}]
          \node[circle, fill=myblue, inner sep=2pt, minimum size=2pt] (a) at (0,0) {a};
          \node[circle, fill=myblue, inner sep=1.4pt, minimum size=2pt] (b) at (2,0) {b};
          \node[circle, fill=myblue, inner sep=2pt, minimum size=2pt] (c) at (2,2) {c};
          \node[circle, draw, inner sep=1.4pt, minimum size=2pt] (d) at (0,2) {d};
        
          \draw[->] (a) -- node[below] {A} (b);
          \draw[->] (b) -- node[right] {B} (c);
          \draw[->] (c) -- node[above left=-2pt] {C} (a);
          \draw[->] (c) -- node[above] {D} (d);
        \end{tikzpicture}
        \vspace{-0.1in}
        \caption{Subgraph isomorphism.}
        \label{fig:preliminary_subiso}
    \end{subfigure}
    \caption{Examples of the isomorphism and subgraph isomorphism, where letters indicate labels.}
    \vspace{-0.2in}
    \label{fig:preliminary_iso_subiso}
\end{wrapfigure}

    Let $\mathcal{G} = (\mathcal{V}_\mathcal{G}, \mathcal{E}_\mathcal{G}, \mathcal{X}_\mathcal{G}, \mathcal{Y}_\mathcal{G})$ be a \textit{graph} with a vertex set $\mathcal{V}_\mathcal{G}$ and each vertex with a different \textit{vertex id}, an edge set $\mathcal{E}_\mathcal{G} \subseteq \mathcal{V}_\mathcal{G} \times \mathcal{V}_\mathcal{G}$,
    a label function $\mathcal{X}_\mathcal{G}$ that maps a vertex to a set of \textit{vertex labels}, and a label function $\mathcal{Y}_\mathcal{G}$ that maps an edge to a set of \textit{edge labels}.
    To simplify the statement, we let $\mathcal{Y}_\mathcal{G}((u, v)) = \emptyset$ (where $\emptyset$ corresponds an empty set) if $(u, v) \not \in \mathcal{E}_\mathcal{G}$.
    A \textit{subgraph} of $\mathcal{G}$, denoted as $\mathcal{G}_S$, is any graph with $\mathcal{V}_{\mathcal{G}_S} \subseteq \mathcal{V}_\mathcal{G}$, $\mathcal{E}_{\mathcal{G}_S} \subseteq \mathcal{E}_\mathcal{G} \cap (\mathcal{V}_{\mathcal{G}_{S}} \times \mathcal{V}_{\mathcal{G}_{S}})$ satisfying $\forall v \in \mathcal{V}_{\mathcal{G}_S}, \mathcal{X}_{\mathcal{G}_S}(v) = \mathcal{X}_{\mathcal{G}}(v)$ and $\forall e \in \mathcal{E}_{\mathcal{G}_S}, \mathcal{Y}_{\mathcal{G}_S}(e) = \mathcal{Y}_{\mathcal{G}}(e)$.
    One of the important properties in graphs is the \textit{isomorphism}.

\begin{definition}[Isomorphism]
    \label{def:iso}
    A graph $\mathcal{G}_1$ is \textit{isomorphic} to a graph $\mathcal{G}_2$ if there is a bijection $f: \mathcal{V}_{\mathcal{G}_1} \rightarrow \mathcal{V}_{\mathcal{G}_2}$ such that:
    \newline\scalebox{0.90}{\parbox{1.11\linewidth}{
    \begin{itemize}[leftmargin=*]
        \item $\forall v \in \mathcal{V}_{\mathcal{G}_1}, \mathcal{X}_{\mathcal{G}_1}(v) = \mathcal{X}_{\mathcal{G}_2}(f(v))$, 
        \item $\forall v' \in \mathcal{V}_{\mathcal{G}_2}, \mathcal{X}_{\mathcal{G}_2}(v') = \mathcal{X}_{\mathcal{G}_1}(f^{-1}(v'))$,
        \item $\forall (u, v) \in \mathcal{E}_{\mathcal{G}_1}, \mathcal{Y}_{\mathcal{G}_1}((u, v)) = \mathcal{Y}_{\mathcal{G}_2}((f(u), f(v)))$,
        \item $\forall (u', v') \in \mathcal{E}_{\mathcal{G}_2}, \mathcal{Y}_{\mathcal{G}_2}((u', v')) = \mathcal{Y}_{\mathcal{G}_1}((f^{-1}(u'), f^{-1}(v')))$.
    \end{itemize}
    }}
\end{definition}
When $\mathcal{G}_1$ and $\mathcal{G}_2$ are isomorphic, we use the notation $\mathcal{G}_1 \simeq \mathcal{G}_2$ to present this and name the function $f$ as an \textit{isomorphism}.
For example, Figure~\ref{fig:preliminary_iso} illustrates two different isomorphisms between the two triangles.
A specific isomorphism $f$ is $\lbrace \rbrace \rightarrow \lbrace \rbrace$ when considering two empty graphs with no vertices.
In addition, the \textit{subgraph isomorphism} is more general.
\begin{definition}[Subgraph isomorphism]
    \label{def:subiso}
    If a subgraph $\mathcal{G}_{1_S}$ of $\mathcal{G}_1$ is isomorphic to a graph $\mathcal{G}_2$ with a bijection $f$,
    we say $\mathcal{G}_{1}$ contains a subgraph isomorphic to $\mathcal{G}_2$ and name $f$ as a \textit{subgraph isomorphism}.
\end{definition}

The problem of subgraph isomorphisms involves two types of subgraphs: node-induced subgraphs and edge-induced subgraphs.
The former one corresponds to induced subgraph definition that requires $\mathcal{E}_{\mathcal{G}_S} = \mathcal{E}_\mathcal{G} \cap (\mathcal{V}_{\mathcal{G}_{S}} \times \mathcal{V}_{\mathcal{G}_{S}})$, while the latter one corresponds to the general definition of subgraphs.
To make it easier to generalize, we assume that all subgraphs mentioned here are edge-induced.


\subsection{Graph Isomorphism Test and Representation Power}
\label{sec:power}
Graph isomorphism tests are used to determine whether two graphs are isomorphic, which are useful in various fields such as computer science, chemistry, and mathematics.
\begin{definition}[Graph Isomorphism Test]
    A \textit{graph isomorphism test} is a function $\chi: \Sigma \times \Sigma \rightarrow \lbrace 0, 1 \rbrace$ that determines whether two graphs are isomorphic, where $\Sigma$ is the graph set.
\end{definition}
Ideally, a perfect graph isomorphism test should be able to distinguish \textbf{all} graph pairs, i.e., $\forall \mathcal{G}_i, \mathcal{G}_j \in \Sigma: \chi(\mathcal{G}_i, \mathcal{G}_j) = 1 \Leftrightarrow \mathcal{G}_i \simeq \mathcal{G}_j$.
The graph isomorphism problem is a well-known computational problem that belongs to the class of NP problems.
Nonetheless, there are heuristic techniques that can solve the graph isomorphism tests for most practical cases.
For example, the \textit{Weisfeiler-Leman algorithm} (WL algorithm)~\cite{weisfeiler1968reduction} is a heuristic test for graph isomorphism that assigns colors to vertices of graphs iteratively:
\newline\scalebox{0.90}{\parbox{1.11\linewidth}{
\begin{align}
    c_{v}^{(t+1)} &= \texttt{Color} \lparen c_{v}^{(t)}, \bm N_{v}^{(t)} \rparen, \nonumber \\
    \bm N_{v}^{(t)} &= \lbrace \texttt{Color} \lparen c_{u}^{(t)}, \lbrace c_{(u, v)}^{(t)} \rbrace \rparen | u \in \mathcal{N}_{v} \rbrace, \nonumber \\
    c_{v}^{(0)} &= \mathcal{X}_{\mathcal{G}}(v),  \nonumber \\
    c_{(u, v)}^{(t)} &= \mathcal{Y}_{\mathcal{G}}\lparen (u, v) \rparen,  \nonumber
\end{align}
}}
where $c_{v}^{(t)}$ is the color of vertex $v$ at the $t$-th iteration, $\mathcal{N}_{v}$ is $v$'s neighbor collection, $\mathcal{X}_{\mathcal{G}}(v)$ is the vertex label of $v$ in graph $\mathcal{G}$, $\mathcal{Y}_{\mathcal{G}}\lparen (u, v) \rparen$ is the edge label of $(u, v)$ in graph $\mathcal{G}$, and $\texttt{Color}$ is a function to assign colors to vertices.
The time complexity of color assignment is $\mathcal{O}(|\mathcal{E}_{\mathcal{G}}|)$ for each iteration.
Given two graphs $\mathcal{G}_i$ and $\mathcal{G}_j$, the WL algorithm refines colors to vertices of $\mathcal{G}_i$ and $\mathcal{G}_j$ in parallel and then compares the resulting colors of vertices in the two graphs.

\begin{figure*}[!t]
    \centering

    \begin{subfigure}[b]{.48\textwidth}
        \centering
        \vspace{-0.15in}
        \includegraphics[height=4.2cm]{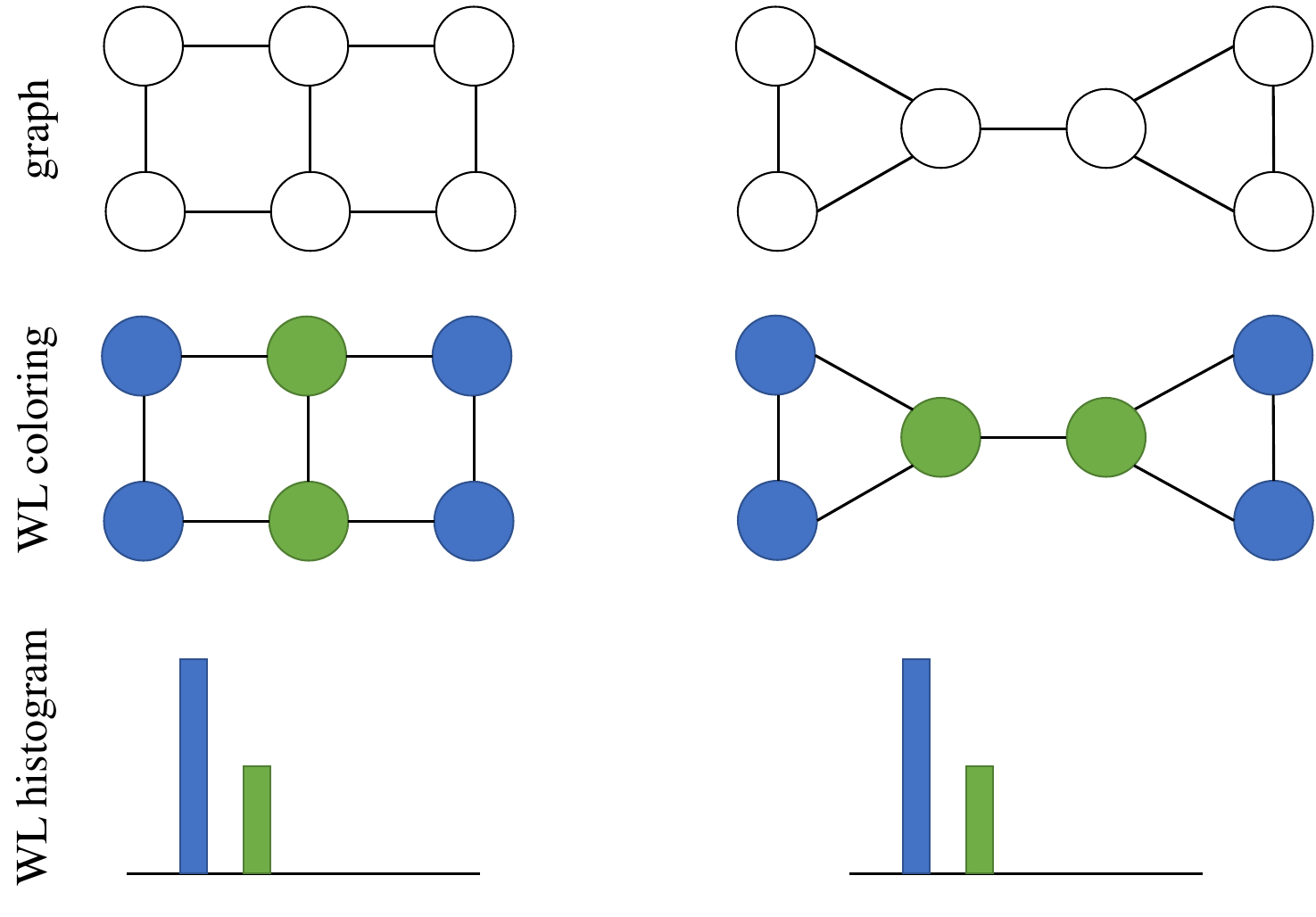}
        \caption{Example of the non-isomorphic graphs with the same WL color histogram.}
    \label{fig:preliminary_wl_color_histogram}
    \end{subfigure}
    \ \ 
    \begin{subfigure}[b]{.48\textwidth}
        \centering
        \vspace{-0.15in}
        \includegraphics[height=4.2cm]
        {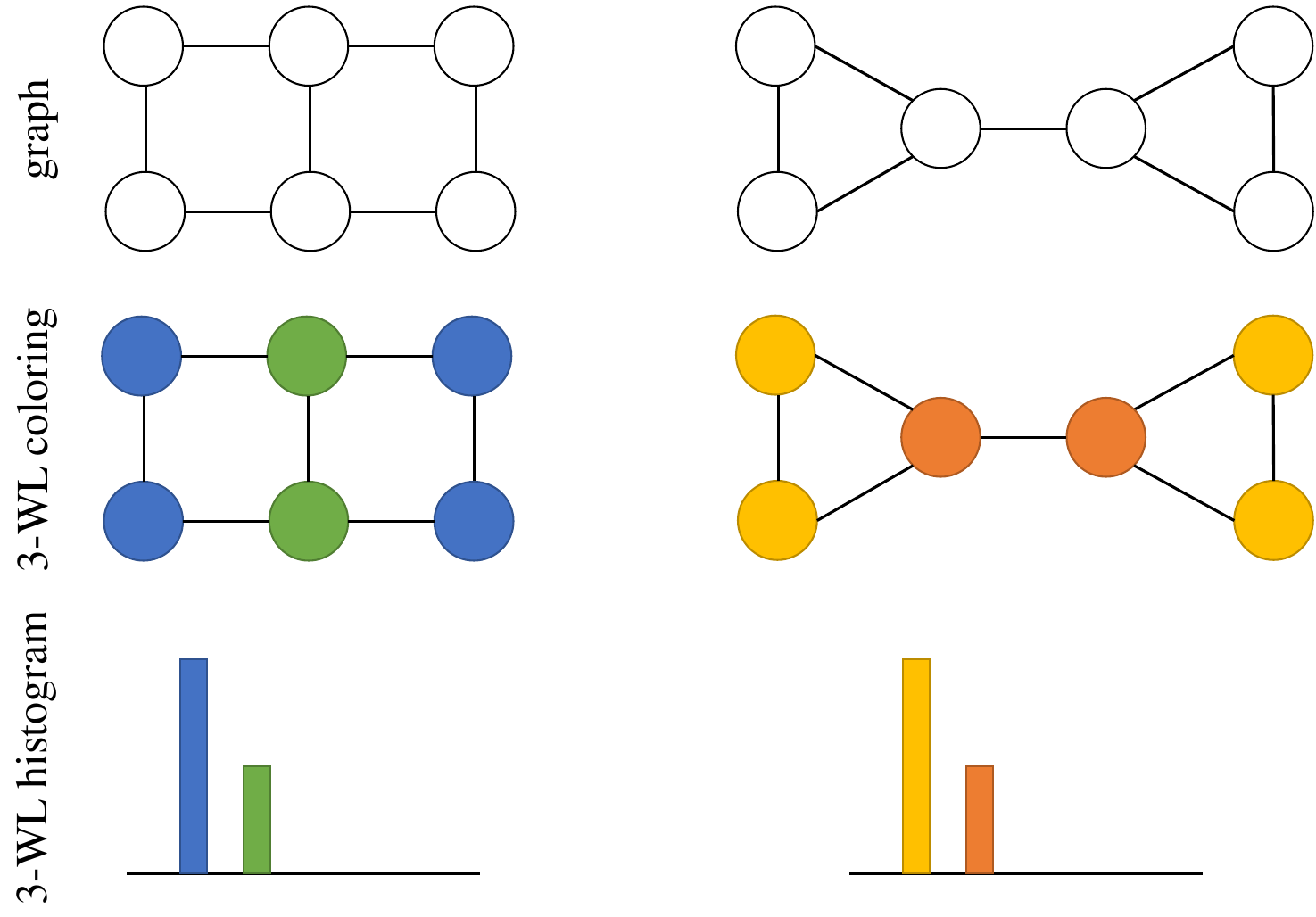}
        \caption{Example of the non-isomorphic graphs with different 3-WL color histograms.}
    \label{fig:preliminary_3wl_color_histogram}
    \end{subfigure}
    \caption{Example of the non-isomorphic graphs with the same WL color histogram but different 3-WL color histograms.}
    \vspace{-0.2in}
    \label{fig:preliminary_color_histogram}
\end{figure*}

The WL algorithm is guaranteed to terminate after a finite number of iterations.
It determines that two graphs are isomorphic only if the colors of vertices in both graphs are the same when the algorithm finishes, here ``same'' refers to having identical color histograms.
However, the WL algorithm is unable to distinguish all non-isomorphic graphs, as demonstrated in Figure~\ref{fig:preliminary_wl_color_histogram}.
Therefore, it is crucial to comprehend the \textbf{representation power} of this algorithm.
What makes the algorithm potent for isomorphism testing is its injective color refinement, which takes into account the neighborhood of each vertex.
The neighbor collection $\mathcal{N}_{v}$ is defined based on the 1-hop neighborhood, meaning that color refinement relies solely on the local information of each vertex.
Hence, an extension of the WL algorithm that considers higher-order interactions between vertices is expected to possess greater power for isomorphism testing, which are called $k$\textit{-dimensional Weisfeiler-Leman algorithm} ($k$-WL)~\cite{cai1992optimal}.
The $k$-WL ($k \geq 2$) iteratively assigns colors to the $k$-tuples of $\mathcal{V}_{\mathcal{G}}^{k}$ as follows:
\newline\scalebox{0.90}{\parbox{1.11\linewidth}{
\begin{align}
    c_{\boldsymbol{v}}^{(t+1)} &= \texttt{Color} \lparen c_{\boldsymbol{v}}^{(t)}, \bm N_{\boldsymbol{v}}^{(t)} \rparen, \nonumber \\
    \bm N_{\boldsymbol{v}}^{(t)} &= \bigcup_{j=1}^{k} \lbrace \texttt{Color} \lparen c_{\boldsymbol{u}}^{(t)}, \lbrace c_{(\boldsymbol{u}, \boldsymbol{v})}^{(t)} \rbrace \rparen | \boldsymbol{u} \in \mathcal{N}_{\boldsymbol{v}}^{j} \rbrace, \nonumber \\
    c_{\boldsymbol{v}}^{(0)} &= \texttt{Color} \lparen \mathcal{X}_{\mathcal{G}}(\boldsymbol{v}[k]), \lbrace \mathcal{X}_{\mathcal{G}}(\boldsymbol{v}) | w \in \boldsymbol{v} - \lbrace \boldsymbol{v}[k] \rbrace \rbrace \rparen , \nonumber \\
    c_{(\boldsymbol{u}, \boldsymbol{v})}^{(t)} &= 
    \begin{cases}
        \mathcal{Y}_{\mathcal{G}}\lparen (u, v) \rparen &\text{ if } u = \boldsymbol{v} - \boldsymbol{u} \text{ and } v = \boldsymbol{u} - \boldsymbol{v}, \\
        \emptyset &\text{  otherwise.}
    \end{cases}, \nonumber 
\end{align}
}}
where $\boldsymbol{v}$ refers to a tuple of $k$ vertices in $\mathcal{V}_{\mathcal{G}}$, $\boldsymbol{v}[j]$ denotes the $j$-th element of the tuple $\boldsymbol{v}$, $\mathcal{N}_{\boldsymbol{v}}^{j}$ ($1 \leq j \leq k$) is the neighbor collection of $\boldsymbol{v}$ in which only the $j$-th elements are different (i.e., $\mathcal{N}_{\boldsymbol{v}}^{j} = \lbrace \boldsymbol{u} | (\forall i \neq j \ \boldsymbol{u}[i] = \boldsymbol{v}[i]) \wedge (\boldsymbol{u}[j] = \boldsymbol{v}[j]) \rbrace$), $\mathcal{X}_{\mathcal{G}}(\boldsymbol{v}[k])$ is the vertex label of $\boldsymbol{v}[k]$ in graph $\mathcal{G}$, $\mathcal{X}_{\mathcal{G}}(\boldsymbol{v})$ is the vertex label of $\boldsymbol{v}$ in graph $\mathcal{G}$, $\mathcal{Y}_{\mathcal{G}}\lparen (u, v) \rparen$ is the edge label of $(u, v)$ in graph $\mathcal{G}$, $\boldsymbol{v} - \boldsymbol{u}$ indicates the difference of two tuples, and $\texttt{Color}$ is a function to assign colors to tuples.
The $k$-WL algorithm is also guaranteed to terminate after a finite number of iterations.
Figure~\ref{fig:preliminary_3wl_color_histogram} demonstrates the same example in Figure~\ref{fig:preliminary_wl_color_histogram} with different 3-WL color histograms, despite having the same WL color histogram.
This suggests that the 3-WL algorithm can distinguish more non-isomorphic graphs than the WL algorithm, which is expected to obtain strictly stronger representation power.
It is worth noting that the complexity of the $k$-WL algorithm increases exponentially because it constructs $|\mathcal{V}_{\mathcal{G}}|^{k}$ tuples and at most $|\mathcal{V}_{\mathcal{G}}|^{2k}$ connections, which are regarded as ``vertices'' and ``edges'' in the $k$-tuple graph.
Thus, the complexity of color assignment becomes $\mathcal{O}(|\mathcal{V}_{\mathcal{G}}|^{2k})$.

\subsection{Graph Kernels}
\label{sec:gk}

However, there is a gap between the power of representation and subgraph isomorphisms.
While the power aims to distinguish non-isomorphic graph pairs, counting subgraph isomorphisms presents a greater challenge, which is a combinatorial problem depending on substructures.
Therefore, our objective is to approximate subgraph isomorphism counting through representation learning and optimization as regression.
The isomorphism test can be seen as a hard indicator function that determines whether an isomorphism exists, which can be extended as a kernel function for subgraph isomprhism counting.
A \textit{kernel function} is designed to measure the similarity between two objects.
\begin{definition}[Kernel]
    A function $k: \Sigma \times \Sigma \rightarrow \mathbb{R}$ is called a \textit{kernel} over an non-empty set $\Sigma$.
\end{definition}
An even more crucial concept is the selection of a kernel function $k$ in a manner that allows for the existence of a \textit{feature map} $h$ from the set $\Sigma$ to a Hilbert space $\mathbb{H}$ equipped with an inner product.
This feature map should satisfy $\forall \mathcal{G}_i, \mathcal{G}_j \in \Sigma: k(\mathcal{G}_i, \mathcal{G}_j) = \langle h(\mathcal{G}_i), h(\mathcal{G}_j) \rangle$. This space $\mathbb{H}$ is referred to as the \textit{feature space}.
A group of kernel functions known as \textit{graph kernels} (GKs) are employed to compute the similarity between two graphs by taking them as input.

Neural networks have been regarded as effective feature extractors and predictors, and sequence and graph neural networks can be aligned with the kernel functions~\cite{lei2017deriving,xu2019how}.
\citet{liu2020neural} proposed a unified end-to-end framework for sequence models and graph models to directly predict the number of subgraph isomorphisms rather than the similarities, further illustrating the practical success.
\section{Beyond the Representation Power Limitation via Implicit Correlations}
\label{sec:method}

Constructing neural networks to directly predict the number of subgraph isomorphisms has been shown effective and efficient~\cite{liu2020neural,liu2021graph,yu2023learning}.
But transforming $\Sigma$ to a limited-dimensional space $\mathbb{H}$ remains challenging in optimization, and it has been shown bounded in theory and practice~\cite{chen2020can,liu2021graph}.
Therefore, we turn to other directions to leverage implicit structure correlations to make predictions.

\subsection{Gram Matrix Construction}
\label{sec:gram}
Given a set of graphs $\mathcal{G}_1, \mathcal{G}_2, \cdots, \mathcal{G}_{D} \in \Sigma$, the \textit{kernel matrix} $\bm{K}$ is defined as:
\newline\scalebox{0.90}{\parbox{1.11\linewidth}{
\begin{align}
    \bm{K} &= 
    \begin{bmatrix}
        k(\mathcal{G}_1, \mathcal{G}_1) & k(\mathcal{G}_1, \mathcal{G}_2) & \cdots & k(\mathcal{G}_1, \mathcal{G}_{D}) \\
        k(\mathcal{G}_2, \mathcal{G}_1) & k(\mathcal{G}_2, \mathcal{G}_2) & \cdots & k(\mathcal{G}_2, \mathcal{G}_{D}) \\
        \vdots & \vdots & \ddots & \vdots \\
        k(\mathcal{G}_{D}, \mathcal{G}_1) & k(\mathcal{G}_{D}, \mathcal{G}_2) & \cdots & k(\mathcal{G}_{D}, \mathcal{G}_{D})
    \end{bmatrix}, \\
    &s.t. \bm{K}_{ij} = k(\mathcal{G}_i, \mathcal{G}_j) = \langle h(\mathcal{G}_i), h(\mathcal{G}_j) \rangle.
\end{align}
}}
The kernel matrix $\bm{K} \in \mathbb{R}^{D \times D}$ is also called the \textit{Gram matrix}.
Different kernel functions emphasize specific structural properties of graphs.
For instance, \textit{Shortest-path Kernel} (SP)~\cite{borgward2005shortest} decomposes graphs into shortest paths and compares graphs according to their shortest paths, such as path lengths and endpoint labels.
Instead, Graphlet Kernels~\cite{shervashidze2009efficient} compute the distribution of small subgraphs (i.e., wedges and triangles) under the assumption that graphs with similar graphlet distributions are highly likely to be similar.
The \textit{Weisfeiler-Leman subtree kernel} (WL kernel) belongs to a family of graph kernels denoted as ($k$-WL)~\cite{shervashidze2011weisfeiler}, where $k$ indicates the element size during the color assignment.

Let's take the WL kernel as an example.
It is a popular graph kernel based on the WL algorithm mentioned in \S~\ref{sec:power} upon 1-hop neighborhood aggregation to assign finite integer labels $\mathbb{S}$, i.e., $h_{\text{KL}}$: $\mathcal{G} \rightarrow \biggm\Vert_{s \in \mathbb{S}} \# \lbrace v \in \mathcal{V}_{\mathcal{G}} | \bm c_{v} = s \rbrace$ such that Color: $v \in \mathcal{V}_{\mathcal{G}} \rightarrow c_{v} \in \mathbb{S}$.
Usually, the convergence of the colors for different graphs occurs in different iterations, so it is hard to determine a specific ``finite'' number of iterations.
Thus, the WL kernel is often manually set to a particular number $T$:
\newline\scalebox{0.90}{\parbox{1.11\linewidth}{
\begin{align}
    k_{\text{WL}}(\mathcal{G}_{i}, \mathcal{G}_{j}) &= \langle h_{\text{WL}}(\mathcal{G}_{i}), h_{\text{WL}}(\mathcal{G}_{j}) \rangle = \biggm\langle \biggm\Vert_{t=0}^{T} \bm C_{\mathcal{V}_{\mathcal{G}_i}}^{(t)} , \biggm\Vert_{t=0}^{T} \bm C_{\mathcal{V}_{\mathcal{G}_j}}^{(t)} \biggm\rangle, \label{eq:wl}
\end{align}
}}
where $\bm C_{\mathcal{V}_{\mathcal{G}_i}}^{(t)}$ is the color histogram (a color vector that counts the number of occurrences of vertex colors) of $\mathcal{G}_{i}$ at iteration $t$, i.e., $\bm C_{\mathcal{V}_{\mathcal{G}_i}}^{(t)} = \biggm\Vert_{s \in \mathbb{S}} \# \lbrace v \in \mathcal{V}_{\mathcal{G}_i} | \bm c_{v}^{(t)} = s \rbrace$.
Note that there is no overlap between the colors at different iterations such that color vectors $\lbrace \bm C_{\mathcal{V}_{\mathcal{G}_i}}^{(t)} | 0 \leq t \leq T \rbrace$ are orthogonal to each other.
Hence, the kernel is efficient in computing by reducing a sum of inner products:
\newline\scalebox{0.90}{\parbox{1.11\linewidth}{
\begin{align}
    k_{\text{WL}}(\mathcal{G}_{i}, \mathcal{G}_{j}) = \sum_{t=0}^{T} {\bm C_{\mathcal{V}_{\mathcal{G}_i}}^{(t)}}^{\top} \bm C_{\mathcal{V}_{\mathcal{G}_j}}^{(t)}. \label{eq:k_wl}
\end{align}
}}
It is worth noting that the Gram matrix does not explicitly maintain the graph structure and substructure information.
But this information can be implicitly captured within the matrix.

The running time for a single color vector is $\mathcal{O}(T \cdot |\mathcal{E}_{\mathcal{G}_i}|)$, and the running time for the dot product is $\mathcal{O}(T \cdot (|\mathcal{V}_{\mathcal{G}_i}| + |\mathcal{V}_{\mathcal{G}_j}|))$.
Therefore, the running time for the WL kernel and the Gram matrix is $\mathcal{O}(T \cdot M \cdot D + T \cdot N \cdot D^2)$, where $N$ and $M$ represent the maximum number of vertices and edges among the 
$D$ graphs, respectively.
For a $k$-WL kernel, the time complexity is $\mathcal{O}(T \cdot N^{2k} \cdot D + T \cdot N \cdot D^2)$.

\subsection{Neighborhood Information in the Hilbert Space}
\label{sec:nie}
Since the graph kernel is a function of the graph representation, the graph structure is expected to be preserved in the Hilbert space.
However, the hash function in the WL kernel family does not capture the neighbors of a node.
For example, $c_u^{(t)}$ and $c_v^{(t)}$ would be different if $u \in \mathcal{V}_{\mathcal{G}i}$ and $v \in \mathcal{V}_{\mathcal{G}_j}$ have different neighbors (more precisely, at least one neighbor is different).
Nevertheless, the subset of neighbors is essential for examining isomorphisms, as the inclusion relation is a necessary condition for subgraph isomorphisms.
We modify the color assignment algorithm in the WL kernel family to incorporate neighborhood information in the Hilbert space or record it in the graph histogram.
The modified WL kernel is called \textit{neighborhood-information-extraction WL kernel} (NIE-WL kernel).
The neighborhood-aware color assignment algorithm is described in Algorithm~\ref{alg:color_assign}.
The only change is the addition of pairwise colors to the color histogram.
This pairwise color depends on the edges and the latest node colors, without affecting the original color assignment.
As a result, the color histogram becomes more expressive, as it can record neighborhood information.

It is clear that the NIE-WL kernel should have the same expressive power as the WL kernel, but the histogram of the NIE-WL kernel records $|\mathcal{V}_{\mathcal{G}}| + |\mathcal{E}_{\mathcal{G}}|$ colors instead of $|\mathcal{V}_{\mathcal{G}}|$ colors.
The additional $|\mathcal{E}_{\mathcal{G}}|$ colors (denoted as $\bm C_{\mathcal{E}_{\mathcal{G}_i}}^{(t)}$ for the $t$-th iteration) can be used in constructing the Gram matrix, where neighborhood information is preserved.
If we decompose the NIE-WL kernel into the WL kernel and the neighborhood-information-extraction kernel (denoted as NIE), we can get:
\newline\scalebox{0.90}{\parbox{1.11\linewidth}{
\begin{align}
    k_{\text{NIE-WL}}(\mathcal{G}_i, \mathcal{G}_j) 
    &= \biggm\langle \biggm\Vert_{t=0}^{T} \bm C_{\mathcal{V}_{\mathcal{G}_i}}^{(t)} || \bm C_{\mathcal{E}_{\mathcal{G}_i}}^{(t)} , \biggm\Vert_{t=0}^{T} \bm C_{\mathcal{V}_{\mathcal{G}_j}}^{(t)} || \bm C_{\mathcal{E}_{\mathcal{G}_j}}^{(t)} \biggm\rangle 
    = \sum_{t=0}^{T} {( \bm C_{\mathcal{V}_{\mathcal{G}_i}}^{(t)} || \bm C_{\mathcal{E}_{\mathcal{G}_i}}^{(t)} )}^{\top} (\bm C_{\mathcal{V}_{\mathcal{G}_j}}^{(t)} || \bm C_{\mathcal{E}_{\mathcal{G}_j}}^{(t)}) \nonumber \\
    &= \sum_{t=0}^{T} {\bm C_{\mathcal{V}_{\mathcal{G}_i}}^{(t)}}^{\top} \bm C_{\mathcal{V}_{\mathcal{G}_j}}^{(t)} + \sum_{t=0}^{T} {\bm C_{\mathcal{E}_{\mathcal{G}_i}}^{(t)}}^{\top} \bm C_{\mathcal{E}_{\mathcal{G}_j}}^{(t)}
    = k_{\text{WL}}(\mathcal{G}_i, \mathcal{G}_j) + k_{\text{NIE}}(\mathcal{G}_i, \mathcal{G}_j). \label{eq:ne_wl}
\end{align}
}}
In other words, the NIE-WL kernel is the linear combination of the WL kernel and the neighborhood-information-extraction kernel.
This also implies that the NIE-WL kernel is a hybrid kernel function, incorporating information beyond feature transformations.

\begin{algorithm}[!t]
    \caption{Neighborhood-aware color assignment algorithm.}
    \label{alg:color_assign}
    {\footnotesize
        \begin{algorithmic}[1]
        \INPUT a directed graph $\mathcal{G} = (\mathcal{V}_{\mathcal{G}}, \mathcal{E}_{\mathcal{G}}, \mathcal{X}_{\mathcal{G}}, \mathcal{Y}_{\mathcal{G}})$, a fixed integer $T$
        \STATE initialize the color of each node $v$ in $\mathcal{V}_{\mathcal{G}}$ as $c_v^{(0)} = \mathcal{X}_{\mathcal{G}}(v)$ and color of each edge $e$ in $\mathcal{E}_{\mathcal{G}}$ as $c_{e} = \mathcal{Y}_{\mathcal{G}}(e)$
        \FOR{iter $t$ from $1$ to $T$}
            \STATE create a new color counter $\mathcal{C}^{(t)}$ and initialize $\mathcal{C}^{(t)} = \emptyset$
            \FOR{each $v$ in $\mathcal{V}_{\mathcal{G}}$}
                \STATE create a color multi-set $\bm{N}_v^{(t)}$ and initialize $\bm{N}_v^{(t)} = \emptyset$
                \FOR{each neighbor $u$ in v's neighbor set $\mathcal{N}_v$}
                    \STATE add the neighbor color $\texttt{Color} \lparen c_u^{(t)}, \lbrace c_{u, v} \rbrace \rparen$ to $\bm{N}_v^{(t)}$
                    \STATE \colorbox{myblue}{calculate the pair-wise color $\texttt{Color} \lparen c_v^{(t)}, \lbrace \texttt{Color} \lparen c_u^{(t)}, \lbrace c_{u, v} \rbrace \rparen \rbrace \rparen$ and record it in the color counter $\mathcal{C}^{(t)}$} \hfill \colorbox{myblue}{//\COMMENT{neighborhood information recording}}
                \ENDFOR
                \STATE calculate the set-wise color $\texttt{Color} \lparen c_v^{(t)}, \bm{N}_v^{(t)} \rparen$, record it in the color counter $\mathcal{C}^{(t)}$, and update $c_v^{(t+1)}$ \hfill //\COMMENT{original color assignment}
            \ENDFOR
        \ENDFOR
        \OUTPUT the graph color histogram $\bigcup_{t=0}^{T} \lbrace \mathcal{C}^{(t)} \rbrace$
        \end{algorithmic}
    }
\end{algorithm}

\subsection{Kernel Tricks}
\label{sec:trick}

Graph kernels are typically characterized as \textit{positive semi-definite kernels}.
Consequently, these kernels $\bm{K}$ possess the \textit{reproducing property}:
\newline\scalebox{0.90}{\parbox{1.11\linewidth}{
\begin{align}
    \theta(\mathcal{G}_i) = \sum_{j=1}^{D} \bm{K}_{ij} \theta(\mathcal{G}_j) = \langle \bm{K}_i , \bm{\theta}\rangle, \label{eq:reproducing}
\end{align}
}}
where $\theta$ is a function belonging to a new feature space $\mathbb{H}'$, and $\bm{\theta} = [\theta(\mathcal{G}_1), \theta(\mathcal{G}_2), \cdots, \theta(\mathcal{G}_{D})]$ is the vectorized representation of $\theta$.
The space $\mathbb{H}'$ is known as the \textit{reproducing kernel Hilbert space} (RKHS) and does not require explicit construction.

Based on the definition of kernels and the reproducing property, a graph $\mathcal{G}_i$ can be represented as $\bm{g}_i$ in the Hilbert space (more precisely, RKHS) according to Eq.~(\ref{eq:reproducing}). 
We then embed the resulting graph representation into another Hilbert space, $\mathbb{H}'$, using another kernel function $k'$.
We consider the following two popular kernel functions.

\paragraph{Polynomial Kernel}

The \textit{polynomial kernel} is defined as $k_{\text{Poly}}(\mathcal{G}_i, \mathcal{G}_j) = (\bm{g}_i^\top \bm{g}_j + 1)^{p}$, where $p \in \mathbb{N}$ is a positive integer.
In practice, explicitly computing the polynomial kernel matrix such that $\bm{K}_{i,j} = k_{\text{Poly}}(\bm{g}_i, \bm{g}_j)$ is infeasible due to the high dimensionality of the Hilbert space.
Instead, we employ the kernel trick to compute the polynomial kernel matrix implicitly:
\newline\scalebox{0.90}{\parbox{1.11\linewidth}{
\begin{align}
k_{\text{Poly}}(\mathcal{G}_i, \mathcal{G}_j) 
= (\bm{g}_i^\top \bm{g}_j + 1)^{p} 
= \sum_{k=0}^{p} \frac{p!}{k!(p-k)!} (\bm{g}_i^\top \bm{g}_j)^{p-k} 1^{k} 
= (\bm{K}_{i, j} + 1)^{p}. \label{eq:poly}
\end{align}
}}

\paragraph{Gaussian Kernel}

A polynomial kernel transforms the graph representation into a higher dimensional space $\binom{|\Sigma| + p}{p}$.
However, the polynomial kernel is sensitive to the parameter $p$, which may result in an overflow issue when $p$ is too large.
A popular kernel function that maps the graph representation into an infinite-dimensional space is named \textit{Gaussian kernel} or \textit{radial basis function kernel}, i.e., $k_{\text{RBF}}(\mathcal{G}_i, \mathcal{G}_j) = \exp(-\frac{\|\bm{g}_i - \bm{g}_j\|^2}{2 \sigma^2})$, where $\sigma$ is a positive real number.
We also have a trick to compute the Gaussian kernel matrix implicitly:
\newline\scalebox{0.90}{\parbox{1.11\linewidth}{
\begin{align}
    k_{\text{RBF}}(\mathcal{G}_i, \mathcal{G}_j) 
    = \exp(-\frac{\|\bm{g}_i - \bm{g}_j\|^2}{2 \sigma^2}) 
    = \exp(-\frac{\bm{g}_i^\top\bm{g}_i - 2 \bm{g}_i^\top\bm{g}_j + \bm{g}_j^\top\bm{g}_j}{2 \sigma^2}) 
    = \exp(-\frac{\bm{K}_{i, i} - 2\bm{K}_{i, j} + \bm{K}_{j, j}}{2 \sigma^2}). \label{eq:rbf}
\end{align}
}}
By employing the aforementioned implicit computation tricks, the kernel transformations become efficient and scalable through matrix operations.
Hybrid kernel functions can be obtained by combining different graph kernels.
For instance, a hybrid kernel function with the WL kernel and RBF kernel is Eq.~(\ref{eq:hybrid}), 
where $\bm{K}_{\text{WL}}$ is the Gram matrix with respect to the WL kernel.
\newline\scalebox{0.90}{\parbox{1.11\linewidth}{
\begin{align}
    k_{\text{WL,RBF}}(\mathcal{G}_i, \mathcal{G}_j) = \exp(-\frac{{\bm{K}_{\text{WL}}}_{i, i} - 2{\bm{K}_{\text{WL}}}_{i, j} + {\bm{K}_{\text{WL}}}_{j, j}}{2 \sigma^2}). \label{eq:hybrid}
\end{align}
}}

\section{Experiment}
\label{sec:experiment}

\subsection{Experimental Setup}

\subsubsection{Evaluation}

We regard subgraph isomorphism counting with graph kernels as a machine learning problem.
Since we model subgraph isomorphism counting as a regression problem with Gram matrices, we use the SVM~\cite{chang2011libsvm,fan2008liblinear} and Ridge~\cite{hoerl1970ridge} implemented in the scikit-learn package.
We evaluate models based on the root mean squared error (RMSE) and the mean absolute error (MAE).
We collect the most popular datasets for graph classification, as graph properties are often determined by substructures within the graph.

In order to obtain meaningful and challenging predictions, we enumerate all vertex label permutations and edge permutations from the 3-stars, triangles, tailed triangles, and chordal cycles.
Furthermore, to improve the quality of the data, we have filtered out patterns with an average frequency of less than 1.0 across the entire dataset.
Detailed settings are reported in Appendix~\ref{appedix:exp_setting}. 
Our approach to graph kernels involves substructures from different levels:

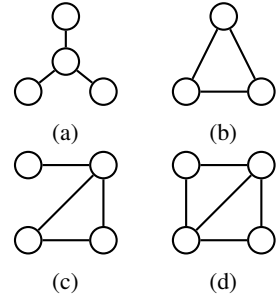
\begin{wrapfigure}{r}{0.3\textwidth}
    \vspace{-0.05in}
    \begin{subfigure}{.48\linewidth}
        \centering
        \begin{tikzpicture}[thick,scale=0.5, every node/.style={scale=1.0}]
           \tikzstyle{every node}=[draw,shape=circle,inner sep=1,minimum size=1.0em];
          \node[circle, draw] (a) at (0,0) {};
          \node[circle, draw] (b) at (-1,-0.8) {};
          \node[circle, draw] (c) at (1,-0.8) {};
          \node[circle, draw] (d) at (0, 1.2) {};
          \draw (a) -- (b);
          \draw (a) -- (c);
          \draw (a) -- (d);
        \end{tikzpicture}
        \caption{}
    \end{subfigure}
    \begin{subfigure}{.48\linewidth}
        \centering
        \begin{tikzpicture}[thick,scale=0.5, every node/.style={scale=1.0}]
           \tikzstyle{every node}=[draw,shape=circle,inner sep=1,minimum size=1.0em];
          \node[circle, draw] (a) at (0,0) {};
          \node[circle, draw] (b) at (2,0) {};
          \node[circle, draw] (c) at (1,2) {};
          \draw (a) -- (b);
          \draw (b) -- (c);
          \draw (c) -- (a);
        \end{tikzpicture}
        \caption{}
    \end{subfigure}
    \\
    \begin{subfigure}{.48\linewidth}
        \centering
        \begin{tikzpicture}[thick,scale=0.5, every node/.style={scale=1.0}]
           \tikzstyle{every node}=[draw,shape=circle,inner sep=1,minimum size=1.0em];
          \node[circle, draw] (a) at (0,0) {};
          \node[circle, draw] (b) at (2,0) {};
          \node[circle, draw] (c) at (2,2) {};
          \node[circle, draw] (d) at (0,2) {};
          \draw (a) -- (b);
          \draw (b) -- (c);
          \draw (c) -- (a);
          \draw (c) -- (d);
        \end{tikzpicture}
        \caption{}
    \end{subfigure}
    \begin{subfigure}{.48\linewidth}
        \centering
        \begin{tikzpicture}[thick,scale=0.5, every node/.style={scale=1.0}]
           \tikzstyle{every node}=[draw,shape=circle,inner sep=1,minimum size=1.0em];
          \node[circle, draw] (a) at (0,0) {};
          \node[circle, draw] (b) at (2,0) {};
          \node[circle, draw] (c) at (2,2) {};
          \node[circle, draw] (d) at (0,2) {};
          \draw (a) -- (b);
          \draw (b) -- (c);
          \draw (c) -- (a);
          \draw (c) -- (d);
          \draw (a) -- (d);
        \end{tikzpicture}
        \caption{}
    \end{subfigure}
    \caption{Four patterns considered in experiments: (a) 3-star, (b) triangle, (c) tailed triangle, and (d) chordal cycle.}
    \label{fig:experiment_patterns}
    \vspace{-0.5in}
\end{wrapfigure}

\ 
\begin{itemize}[leftmargin=*]
    \item \textbf{Paths}: Shortest-path Kernel~\cite{borgward2005shortest} decomposes graphs into shortest paths and compares these paths.
    \item \textbf{Wedges and triangles}: Graphlet Kernels~\cite{shervashidze2009efficient} compute the distribution of graphlets of size 3, which consist of wedges and triangles.
    \item \textbf{Whole graphs}: The Weisfeiler-Leman Optimal Assignment Kernel (WLOA)\cite{kriege2016on} improves the performance of the WL kernel by capitalizing on the theory of valid assignment kernels.
    \item \textbf{High-order neighborhood}: $k$-WL Kernels\cite{shervashidze2011weisfeiler} measure the histogram of $k$-combinations by assigning colors to $k$-tuples, while $\delta$-$k$ WL kernels record the number of $k$-tuples with the same color. $\delta$-$k$-LWL and $\delta$-$k$-LWL$^{+}$ focus on local structures of the graph instead of the whole graph.
\end{itemize}

We also apply the polynomial and Gaussian kernel tricks to the above kernels.
Besides, we incorporate the neighborhood-aware color assignment to $k$-WL Kernels and their variants.

\subsubsection{Efficient Implementation}

\begin{figure*}[h!]
    \vspace{-0.1in}
    \centering
    \includegraphics[width=0.6\linewidth]{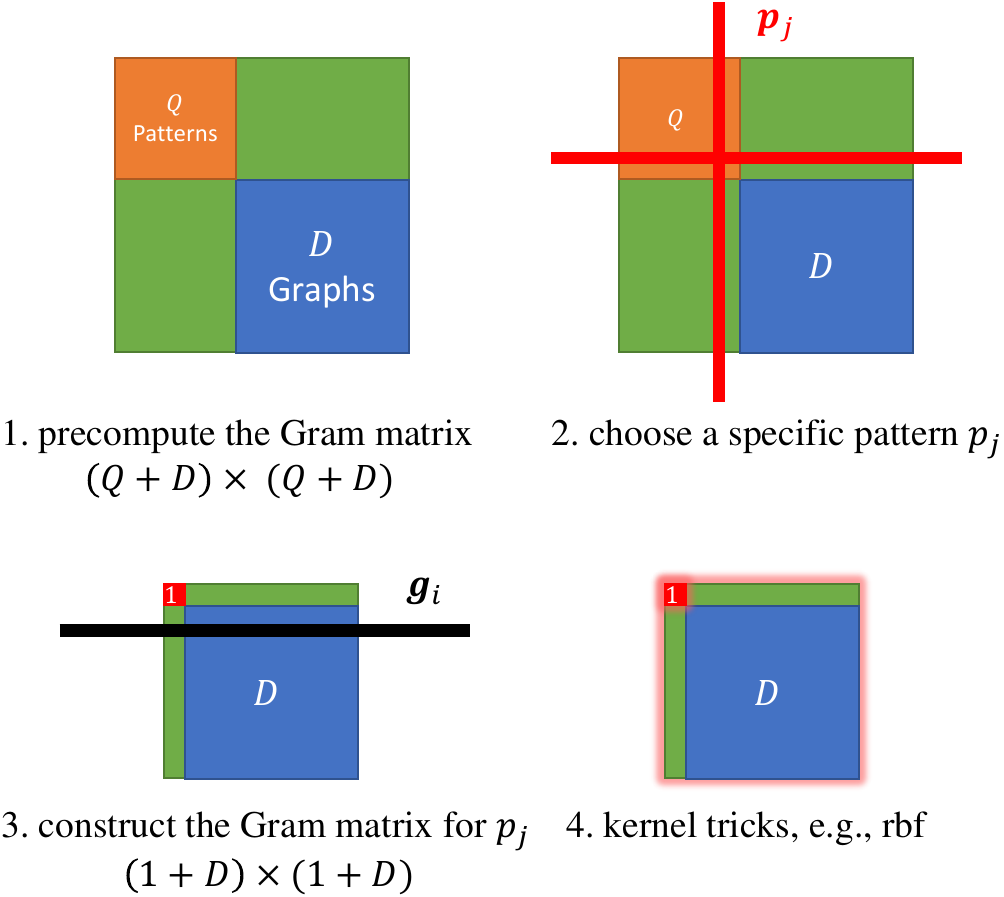}
    \caption{Efficient implementation.}
    \label{fig:kernel_implementation}
\end{figure*}

Graph kernels are implemented in C++ with C++11 features and the -O2 flag.
In addition to the technical aspects, we also focus on training efficiency.
The simplest input for regressors is the original graph kernel matrix of size $D \times D$.
However, this kernel matrix only contains graphs in the entire dataset, without any information about patterns.
It is necessary to incorporate pattern structure information during training.
Assuming we have $Q$ patterns, we need to repeatedly construct $Q$ kernel matrices of size $(1 + D) \times (1 + D)$.
In fact, the $D \times D$ submatrix is the same for all $Q$ kernel matrices because it is irrelevant to the patterns.
Therefore, it is recommended that we construct a matrix of size $(Q + D) \times (Q + D)$ only once and then repeatedly slice the submatrix to obtain information about the $D$ graphs and the specific pattern for prediction.

\subsection{Empirical Results}

\subsubsection{SVM vs. Ridge}

We begin by comparing the performance of SVM and Ridge (precisely, Kernel Ridge) regression without kernel tricks on subgraph isomorphism counting.
This is a common practice to evaluate the performance of these two regressors.
The results are shown in Figures~\ref{fig:svm_linear} and ~\ref{fig:ridge_linear}, and we can see that the performance of the two regressors is comparable, with Ridge performing slightly better.
There are two main reasons for this.
First, Ridge is solved by Cholesky factorization in the closed form, which typically achieves better convergence than the iterative optimization of SVM.
Second, the objective of Ridge is to minimize the sum of squared errors, which is more straightforward and suitable for the regression task than the $\epsilon$-insensitive loss function of SVM.
Given the superior performance of Ridge, we only report the results of Ridge in the following experiments to save space.

\subsubsection{Kernel Tricks for Implicit Transform}
In addition, we also observe an increase in errors with the polynomial kernel trick from Figure~\ref{fig:svm_ridge}.
The number of matched subgraphs is typically small but can be very large for specific structures, such as complete graphs or cliques.
The polynomial kernel trick can easily lead to fluctuations due to extreme values, resulting in performance fluctuations and even overflow.

\begin{figure*}[t]
    \centering
    \begin{subfigure}[b]{0.24\linewidth}
        \includegraphics[width=\linewidth]{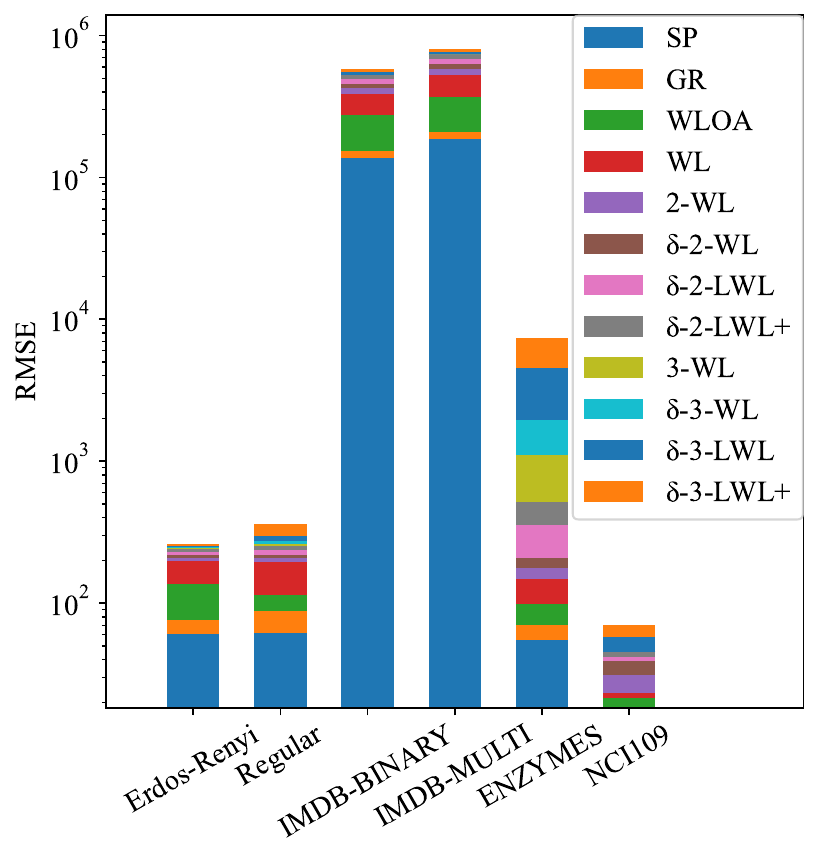}
        \vspace{-0.2in}
        \caption{SVM, Linear.}
        \label{fig:svm_linear}
    \end{subfigure}
    \begin{subfigure}[b]{0.24\linewidth}
        \includegraphics[width=\linewidth]{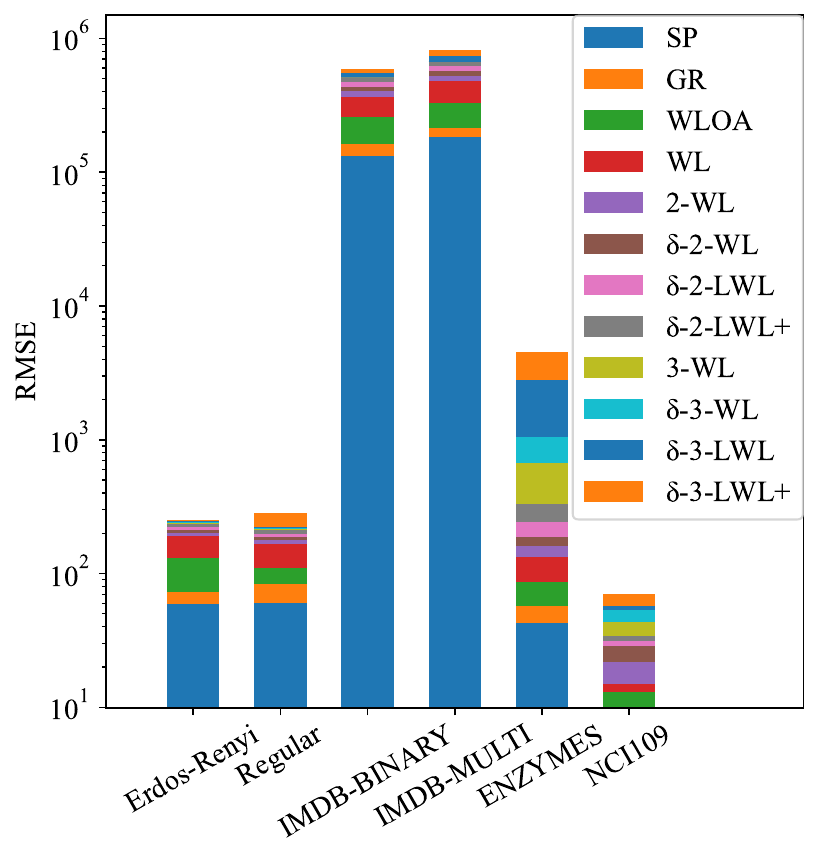}
        \vspace{-0.2in}
        \caption{Ridge, Linear.}
        \label{fig:ridge_linear}
    \end{subfigure}
    \begin{subfigure}[b]{0.24\linewidth}
        \includegraphics[width=\linewidth]{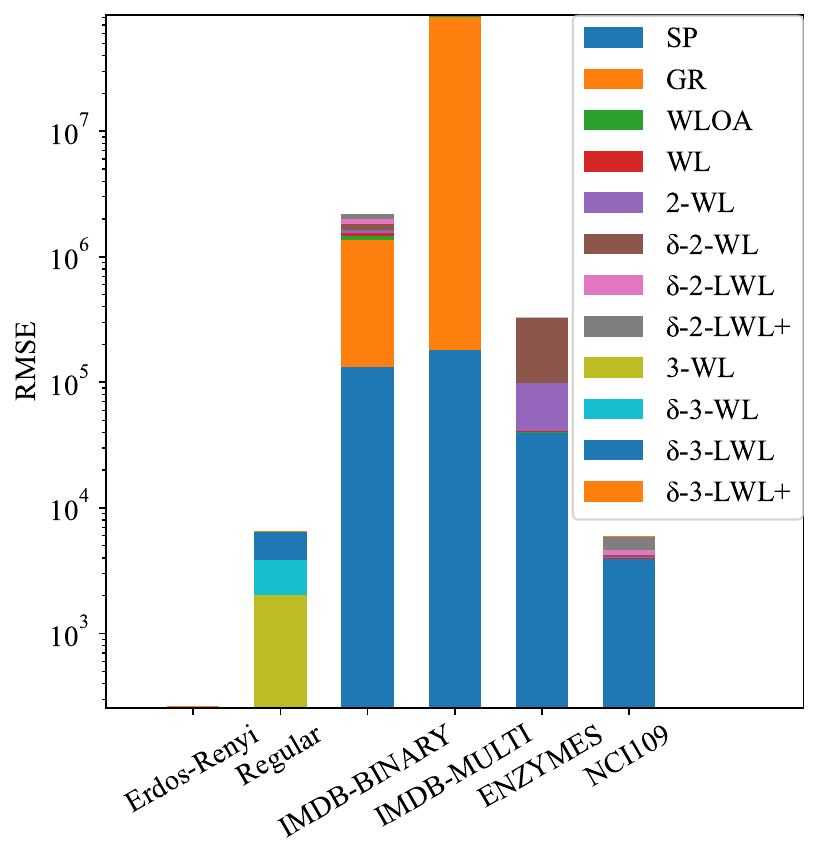}
        \vspace{-0.2in}
        \caption{Ridge, Poly.}
        \label{fig:ridge_poly}
    \end{subfigure}
    \begin{subfigure}[b]{0.24\linewidth}
        \includegraphics[width=\linewidth]{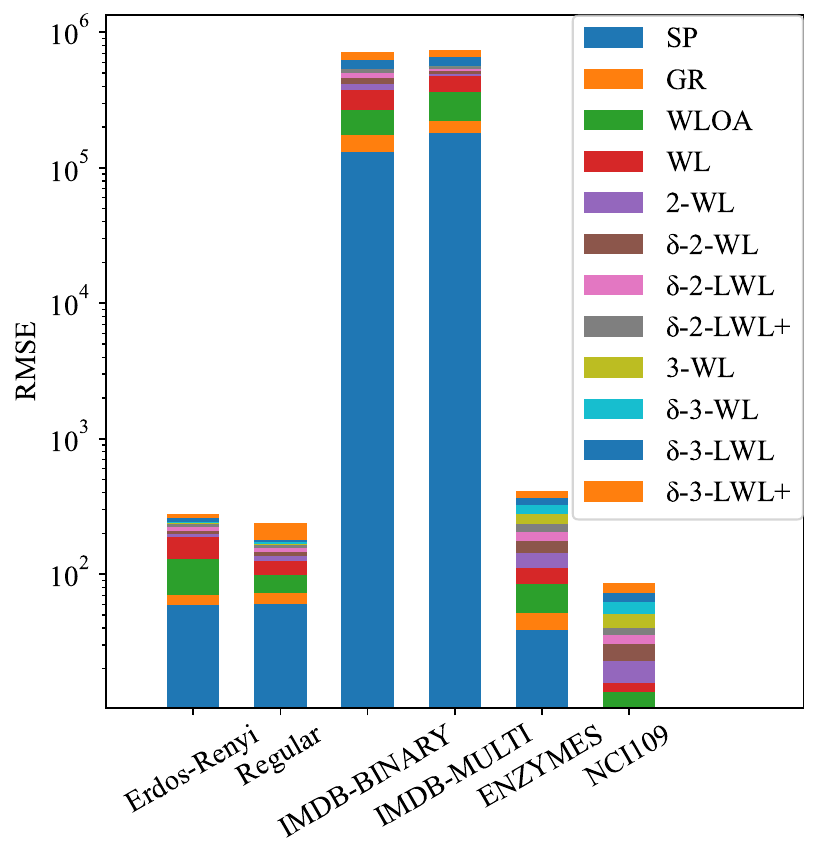}
        \vspace{-0.2in}
        \caption{Ridge, RBF.}
        \label{fig:ridge_rbf}
    \end{subfigure}
    \setcounter{figure}{{\value{figure}-1}}
    \vspace{-0.05in}
    \caption{Illustration on subgraph isomorphism counting, where the out-of-memory (``OOM'') is regarded as zero in the plots.}
    \label{fig:svm_ridge}
\end{figure*}

\begin{table*}[h!]
    \scriptsize 
    \centering 
    \setlength\tabcolsep{2.3pt} 
    \vspace{-0.1in}
    \caption{Performance on subgraph isomorphism counting, where $k$-WL$\ddag$ represents the best model in the kernel family, and the best and second best are marked in \textcolor{red}{red} and \textcolor{blue}{blue}, respectively.} 
    \label{tab:exp} 
    \begin{tabular}{l|cc|cc|cc|cc|cc|cc} 
        \toprule 
        \multicolumn{1}{c|}{\multirow{3}{*}{Models}} & \multicolumn{8}{c|}{\textbf{Homogeneous}} & \multicolumn{4}{c}{\textbf{Heterogeneous}} \\
        & \multicolumn{2}{c|}{\textit{Erd\H{o}s-Renyi}} & \multicolumn{2}{c|}{\textit{Regular}} & \multicolumn{2}{c|}{IMDB-BINARY} & \multicolumn{2}{c|}{IMDB-MULTI} & \multicolumn{2}{c|}{\textit{ENZYMES}} & \multicolumn{2}{c}{\textit{NCI109}} \\
        & RMSE & MAE & RMSE & MAE & RMSE & MAE & RMSE & MAE & RMSE & MAE & RMSE & MAE \\
        \midrule 
        \multirow{1}{*}{Zero} & 92.532 & 51.655 & 198.218 & 121.647 & 138262.003 & 37041.171 & 185665.874 & 33063.770 & 64.658 & 25.110 & 16.882 & 7.703 \\
        \multirow{1}{*}{Avg} & 121.388 & 131.007 & 156.515 & 127.211 & 133228.554 & 54178.671 & 182717.385 & 53398.301 & 59.589 & 31.577 & 14.997 & 8.622 \\
        \multirow{1}{*}{CNN} & 20.386 & 13.316 & 37.192 & 27.268 & 4808.156 & 1570.293 & 4185.090 & 1523.731 & 16.752 & 7.720 & 3.096 & 1.504 \\
        \multirow{1}{*}{LSTM} & 14.561 & 9.949 & 14.169 & 10.064 & 10596.339 & 2418.997 & 10755.528 & 1925.363 & 20.211 & 8.841 & 4.467 & 2.234 \\
        \multirow{1}{*}{TXL} & 10.861 & 7.105 & 15.263 & 10.721 & 15369.186 & 3170.290 & 19706.248 & 3737.862 & 25.912 & 11.284 & 5.482 & 2.823 \\
        \multirow{1}{*}{RGCN} & 9.386 & 5.829 & 14.789 & 9.772 & 46074.355 & 13498.414 & 69485.242 & 12137.598 & 23.715 & 11.407 & \bf\color{blue} 1.217 & 0.622 \\
        \multirow{1}{*}{RGIN} & 6.063 & 3.712 & 13.554 & 8.580 & 31058.764 & 6445.103 & 26546.882 & 4508.339 & \bf\color{red} 8.119 & \bf\color{red} 3.783 & \bf\color{red} 0.511 & \bf\color{red} 0.292 \\
        \multirow{1}{*}{CompGCN} & 6.706 & 4.274 & 14.174 & 9.685 & 32145.713 & 8576.071 & 26523.880 & 7745.702 & 14.985 & 6.438 & 1.271 & 0.587 \\
        \hline 
        \multicolumn{13}{c}{\textbf{Ridge, Linear}} \\ 
        \hline 
        SP & 58.721 & 34.606 & 60.375 & 41.110 & 131672.705 & 56058.593 & 181794.702 & 54604.564 & 43.007 & 14.422 & 4.824 & 2.268 \\
        GR & 14.067 & 7.220 & 23.775 & 12.172 & 30527.764 & 7894.568 & 30980.135 & 6054.027 & 14.557 & 5.595 & 5.066 & 2.066 \\
        WLOA & 58.719 & 34.635 & 25.905 & 17.003 & 96887.226 & 28849.659 & 117828.698 & 25808.362 & 28.911 & 11.807 & 3.142 & 1.142 \\
        WL & 58.719 & 34.635 & 56.045 & 33.891 & 107500.276 & 41523.359 & 147822.358 & 49244.753 & 46.466 & 14.920 & 1.896 & 0.746 \\
        2-WL$\ddag$ & 9.921 & 4.164 & 8.751 & 5.663 & 33336.019 & 9161.265 & 47075.541 & 13751.520 & 26.903 & 10.079 & 2.584 & 1.068 \\
        3-WL$\ddag$ & \bf\color{red} 4.096 & \bf\color{red} 1.833 & \bf\color{red} 3.975 & \bf\color{red} 2.277 & 39237.071 & 7240.730 & 76218.966 & 9022.754 & 335.940 & 13.790 & 3.872 & 1.375 \\
        \hline 
        \multicolumn{13}{c}{\textbf{Ridge, Linear, NIE}} \\ 
        \hline 
        WLOA & 58.719 & 34.635 & 25.905 & 17.003 & 33625.086 & 6009.372 & 20858.288 & 2822.391 & 23.478 & 10.037 & 3.203 & 1.133 \\
        WL & 58.719 & 34.635 & 56.045 & 33.891 & 66414.032 & 17502.328 & 70013.402 & 13266.318 & 20.971 & 8.672 & 1.772 & 0.704 \\
        2-WL$\ddag$ & 9.921 & 4.164 & 8.751 & 5.663 & 14914.025 & 3671.681 & 37676.903 & 9930.398 & 97.024 & 7.191 & 1.259 & \bf\color{blue} 0.539 \\
        3-WL$\ddag$ & \bf\color{red} 4.096 & \bf\color{red} 1.833 & \bf\color{red} 3.975 & \bf\color{red} 2.277 & \bf\color{blue} 1808.841 & \bf\color{blue} 264.480 & \bf\color{blue} 1346.608 & \bf\color{blue} 123.394 & 380.480 & 19.073 & OOM & OOM \\
        \hline 
        \multicolumn{13}{c}{\textbf{Ridge, RBF}} \\ 
        \hline 
        SP & 58.721 & 34.606 & 60.375 & 41.110 & 131672.705 & 56058.593 & 181794.702 & 54604.564 & 38.945 & 14.712 & 5.474 & 2.224 \\
        GR & 11.670 & 5.663 & 12.488 & 5.012 & 42387.021 & 5110.985 & 41171.761 & 4831.495 & \bf\color{blue} 12.883 & \bf\color{blue} 5.073 & 4.804 & 1.944 \\
        WLOA & 58.719 & 34.635 & 25.906 & 17.002 & 92733.105 & 28242.033 & 137300.092 & 34067.513 & 32.827 & 12.230 & 3.215 & 1.261 \\
        WL & 58.719 & 34.635 & 25.905 & 17.003 & 109418.159 & 32350.523 & 112515.690 & 25035.268 & 26.313 & 10.933 & 2.227 & 0.837 \\
        2-WL$\ddag$ & 10.500 & 4.630 & 8.495 & 5.634 & 40412.745 & 5351.789 & 21910.109 & 2982.532 & 29.560 & 11.878 & 5.001 & 1.799 \\
        3-WL$\ddag$ & \bf\color{blue} 4.896 & \bf\color{blue} 2.536 & \bf\color{blue} 4.567 & \bf\color{blue} 2.745 & 89532.736 & 21918.757 & 91445.323 & 17703.656 & 43.908 & 18.509 & 10.925 & 5.320 \\
        \hline 
        \multicolumn{13}{c}{\textbf{Ridge, RBF, NIE}} \\ 
        \hline 
        WLOA & 58.719 & 34.635 & 25.906 & 17.002 & 31409.659 & 6644.798 & 19456.664 & 3892.678 & 24.429 & 10.354 & 3.163 & 1.189 \\
        WL & 58.719 & 34.635 & 25.905 & 17.003 & 48568.177 & 17533.158 & 71434.770 & 20472.124 & 23.155 & 9.302 & 2.026 & 0.805 \\
        2-WL$\ddag$ & 10.500 & 4.630 & 8.495 & 5.634 & 15241.302 & 3289.949 & 30093.401 & 6593.717 & 33.838 & 13.947 & 6.619 & 2.807 \\
        3-WL$\ddag$ & \bf\color{blue} 4.896 & \bf\color{blue} 2.536 & \bf\color{blue} 4.567 & \bf\color{blue} 2.745 & \bf\color{red} 757.736 & \bf\color{red} 148.417 & \bf\color{red} 833.037 & \bf\color{red} 75.286 & 43.918 & 18.491 & OOM & OOM \\
        \bottomrule 
    \end{tabular} 
    \vspace{-0.1in}
\end{table*}

\subsubsection{Effectiveness of Neighborhood Information Extraction}

Explicit neighborhood information extraction (NIE) is a crucial component for handling homogeneous data by providing edge colors.
However, this method is not as beneficial when applied to synthetic \textit{Erd\H{o}s-Renyi} and \textit{Regular} datasets because the uniform distribution of neighborhoods results in uniform distributions of edge colors.
As demonstrated in Table~\ref{tab:exp}, incorporating NIE consistently enhances the performance of both linear and RBF kernels.

Overall, the RBF kernel combined with NIE proves to be more effective for homogeneous data, while the linear kernel is substantially improved when applied to heterogeneous data.
The most significant enhancements are observed on the highly challenging \textit{IMDB-BINARY} and \textit{IMDB-MULTI} datasets, where the RMSE is dramatically reduced from 30,527.764 to 757.736 and from 21,910.109 to 833.037, respectively.
When compared to naive baselines that predict either zeros or the training sets' expectations, the RMSE is diminished to a mere 0.5\%.
In addition, some kernel methods, such as GR and the 2-WL family, can provide the same or even more reliable predictions as neural methods.
Moreover, NIE attains state-of-the-art performance on homogeneous data, with a relative gain of 72.1\% compared with the best neural networks.
As for the remaining two heterogeneous datasets, neighborhood information is still not comparable to the GR kernel in \textit{ENZYMES}.
This observation is aligned with the performance of CompGCN~\cite{vashishth2020composition}, where such the node-edge composition may hurt the structure presentation.
RGIN~\cite{liu2020neural} significantly outperforms graph kernels, indicating the future direction of advanced subset representations.



\section{Conclusion, Limitations, and Future Work}
\label{sec:conclusion}
We are the first to concentrate on the representation of patterns and subgraphs by utilizing a variety of graph kernels to tackle the challenge of subgraph isomorphism counting.
While most graph kernels are designed for substructures, their application in approximating subgraph isomorphism counting is not straightforward.
Instead, we propose constructing the Gram matrix to leverage implicit correlations.
Experimental results demonstrate the effectiveness of graph kernels and kernel tricks. Additionally, neighborhood information extraction (NIE) could relieve overfitting by additional pair-wise histograms and obtain significant improvement in challenging datasets.

Upon analyzing the differences between the graphlet kernel and the NIE-WL hybrid kernel, we observe that high-order topologies like triangles and wedges may possess greater power than the first-order.
However, we also note that 3-WL-family kernels perform poorly on heterogeneous data.
On the other hand, RGIN exceeds graph kernels with the help of superior subset representations.
These significant findings serve as a foundation for further research and advancements in the field of graph kernels, as well as other representation learning methods like graph neural networks.

\clearpage
\newpage
\bibliographystyle{plainnat}
\bibliography{neurips_2024}

\begin{thebibliography}{56}
\providecommand{\natexlab}[1]{#1}
\providecommand{\url}[1]{\texttt{#1}}
\expandafter\ifx\csname urlstyle\endcsname\relax
  \providecommand{\doi}[1]{doi: #1}\else
  \providecommand{\doi}{doi: \begingroup \urlstyle{rm}\Url}\fi

\bibitem[Alon et~al.(1995)Alon, Yuster, and Zwick]{alon1995color}
Noga Alon, Raphael Yuster, and Uri Zwick.
\newblock Color-coding.
\newblock \emph{Journal of the ACM}, 42\penalty0 (4):\penalty0 844--856, 1995.

\bibitem[Alon et~al.(2008)Alon, Dao, Hajirasouliha, Hormozdiari, and
  Sahinalp]{alon2008biomolecular}
Noga Alon, Phuong Dao, Iman Hajirasouliha, Fereydoun Hormozdiari, and S~Cenk
  Sahinalp.
\newblock Biomolecular network motif counting and discovery by color coding.
\newblock \emph{Bioinformatics}, 24\penalty0 (13):\penalty0 i241--i249, 2008.

\bibitem[Bause and Kriege(2022)]{bause2022lgradual}
Franka Bause and Nils~Morten Kriege.
\newblock Gradual weisfeiler-leman: Slow and steady wins the race.
\newblock In \emph{{LoG}}, volume 198 of \emph{Proceedings of Machine Learning
  Research}, page~20. {PMLR}, 2022.

\bibitem[Borgwardt and Kriegel(2005)]{borgward2005shortest}
Karsten~M. Borgwardt and Hans{-}Peter Kriegel.
\newblock Shortest-path kernels on graphs.
\newblock In \emph{ICDM}, pages 74--81, 2005.

\bibitem[Bressan et~al.(2019)Bressan, Leucci, and Panconesi]{bressan2019motivo}
Marco Bressan, Stefano Leucci, and Alessandro Panconesi.
\newblock Motivo: fast motif counting via succinct color coding and adaptive
  sampling.
\newblock \emph{PVLDB}, 12\penalty0 (11):\penalty0 1651--1663, 2019.

\bibitem[Cai et~al.(1992)Cai, F{\"u}rer, and Immerman]{cai1992optimal}
Jin-Yi Cai, Martin F{\"u}rer, and Neil Immerman.
\newblock An optimal lower bound on the number of variables for graph
  identifications.
\newblock \emph{Combinatorica}, 12\penalty0 (4):\penalty0 389--410, 1992.

\bibitem[Carletti et~al.(2018)Carletti, Foggia, Saggese, and
  Vento]{carletti2018challenging}
Vincenzo Carletti, Pasquale Foggia, Alessia Saggese, and Mario Vento.
\newblock Challenging the time complexity of exact subgraph isomorphism for
  huge and dense graphs with {VF3}.
\newblock \emph{IEEE Transactions on Pattern Analysis and Machine
  Intelligence}, 40\penalty0 (4):\penalty0 804--818, 2018.

\bibitem[Chang and Lin(2011)]{chang2011libsvm}
Chih-Chung Chang and Chih-Jen Lin.
\newblock Libsvm: a library for support vector machines.
\newblock \emph{Transactions on Intelligent Systems and Technology}, 2\penalty0
  (3):\penalty0 1--27, 2011.

\bibitem[Chen et~al.(2020)Chen, Chen, Villar, and Bruna]{chen2020can}
Zhengdao Chen, Lei Chen, Soledad Villar, and Joan Bruna.
\newblock Can graph neural networks count substructures?
\newblock In \emph{NeurIPS}, volume~33, pages 10383--10395, 2020.

\bibitem[Cheng et~al.(2007)Cheng, Ke, Ng, and Lu]{cheng2007fgindex}
James Cheng, Yiping Ke, Wilfred Ng, and An~Lu.
\newblock Fg-index: towards verification-free query processing on graph
  databases.
\newblock In \emph{{SIGMOD}}, pages 857--872, 2007.

\bibitem[Claris{\'o} and Cabot(2018)]{clariso2018applying}
Robert Claris{\'o} and Jordi Cabot.
\newblock Applying graph kernels to model-driven engineering problems.
\newblock In \emph{{MASES}}, pages 1--5, 2018.

\bibitem[Cordella et~al.(2004{\natexlab{a}})Cordella, Foggia, Sansone, and
  Vento]{cordella2004a}
Luigi~P. Cordella, Pasquale Foggia, Carlo Sansone, and Mario Vento.
\newblock A (sub)graph isomorphism algorithm for matching large graphs.
\newblock \emph{IEEE Transactions on Pattern Analysis and Machine
  Intelligence}, 26\penalty0 (10):\penalty0 1367--1372, 2004{\natexlab{a}}.

\bibitem[Cordella et~al.(2004{\natexlab{b}})Cordella, Foggia, Sansone, and
  Vento]{vento2004a}
Luigi~P. Cordella, Pasquale Foggia, Carlo Sansone, and Mario Vento.
\newblock A (sub)graph isomorphism algorithm for matching large graphs.
\newblock \emph{Transactions on Pattern Analysis and Machine Intelligence},
  26\penalty0 (10):\penalty0 1367--1372, 2004{\natexlab{b}}.

\bibitem[Dai et~al.(2019)Dai, Yang, Yang, Carbonell, Le, and
  Salakhutdinov]{transformerxl2019dai}
Zihang Dai, Zhilin Yang, Yiming Yang, Jaime~G. Carbonell, Quoc~Viet Le, and
  Ruslan Salakhutdinov.
\newblock Transformer-xl: Attentive language models beyond a fixed-length
  context.
\newblock In \emph{ACL}, pages 2978--2988, 2019.

\bibitem[Fan et~al.(2008)Fan, Chang, Hsieh, Wang, and Lin]{fan2008liblinear}
Rong-En Fan, Kai-Wei Chang, Cho-Jui Hsieh, Xiang-Rui Wang, and Chih-Jen Lin.
\newblock Liblinear: A library for large linear classification.
\newblock \emph{Journal of Machine Learning Research}, 9:\penalty0 1871--1874,
  2008.

\bibitem[Gilmer et~al.(2017)Gilmer, Schoenholz, Riley, Vinyals, and
  Dahl]{gilmer2017neural}
Justin Gilmer, Samuel~S. Schoenholz, Patrick~F. Riley, Oriol Vinyals, and
  George~E. Dahl.
\newblock Neural message passing for quantum chemistry.
\newblock In \emph{ICML}, volume~70, pages 1263--1272, 2017.

\bibitem[Giugno and Shasha(2002)]{giugno2002graphgrep}
Rosalba Giugno and Dennis~E. Shasha.
\newblock Graphgrep: {A} fast and universal method for querying graphs.
\newblock In \emph{{ICPR}}, pages 112--115, 2002.

\bibitem[Glava{\v{s}} and {\v{S}}najder(2013)]{glavavs2013recognizing}
Goran Glava{\v{s}} and Jan {\v{S}}najder.
\newblock Recognizing identical events with graph kernels.
\newblock In \emph{{ACL}}, pages 797--803, 2013.

\bibitem[Han et~al.(2013)Han, Lee, and Lee]{han2013turbo}
Wook{-}Shin Han, Jinsoo Lee, and Jeong{-}Hoon Lee.
\newblock Turbo\({}_{\mbox{iso}}\): towards ultrafast and robust subgraph
  isomorphism search in large graph databases.
\newblock In \emph{SIGMOD}, pages 337--348, 2013.

\bibitem[He and Singh(2008)]{he2008graphs}
Huahai He and Ambuj~K. Singh.
\newblock Graphs-at-a-time: query language and access methods for graph
  databases.
\newblock In \emph{SIGMOD}, pages 405--418, 2008.

\bibitem[Hochreiter and Schmidhuber(1997)]{hochreiter1997long}
Sepp Hochreiter and J{\"{u}}rgen Schmidhuber.
\newblock Long short-term memory.
\newblock \emph{Neural Computation}, 9\penalty0 (8):\penalty0 1735--1780, 1997.

\bibitem[Hoerl and Kennard(1970)]{hoerl1970ridge}
Arthur~E Hoerl and Robert~W Kennard.
\newblock Ridge regression: Biased estimation for nonorthogonal problems.
\newblock \emph{Technometrics}, 12\penalty0 (1):\penalty0 55--67, 1970.

\bibitem[Huang et~al.(2016)Huang, Zheng, Cheng, Sun, Mamoulis, and
  Li]{huang2016meta}
Zhipeng Huang, Yudian Zheng, Reynold Cheng, Yizhou Sun, Nikos Mamoulis, and
  Xiang Li.
\newblock Meta structure: Computing relevance in large heterogeneous
  information networks.
\newblock In \emph{SIGKDD}, pages 1595--1604, 2016.

\bibitem[Jha et~al.(2015)Jha, Seshadhri, and Pinar]{jha2015path}
Madhav Jha, C.~Seshadhri, and Ali Pinar.
\newblock Path sampling: {A} fast and provable method for estimating 4-vertex
  subgraph counts.
\newblock In \emph{WWW}, pages 495--505, 2015.

\bibitem[Jie et~al.(2014)Jie, Zhang, Wee, and Shen]{jie2014topological}
Biao Jie, Daoqiang Zhang, Chong-Yaw Wee, and Dinggang Shen.
\newblock Topological graph kernel on multiple thresholded functional
  connectivity networks for mild cognitive impairment classification.
\newblock \emph{Human brain mapping}, 35\penalty0 (7):\penalty0 2876--2897,
  2014.

\bibitem[Kashima and Inokuchi(2002)]{kashima2002kernels}
Hisashi Kashima and Akihiro Inokuchi.
\newblock Kernels for graph classification.
\newblock In \emph{ICDM workshop}, volume 2002, 2002.

\bibitem[Kim(2014)]{kim2014convolutional}
Yoon Kim.
\newblock Convolutional neural networks for sentence classification.
\newblock In \emph{EMNLP}, pages 1746--1751, 2014.

\bibitem[Klein et~al.(2011)Klein, Kriege, and Mutzel]{klein2011ctindex}
Karsten Klein, Nils~M. Kriege, and Petra Mutzel.
\newblock {CT-index}: Fingerprint-based graph indexing combining cycles and
  trees.
\newblock In \emph{{ICDE}}, pages 1115--1126, 2011.

\bibitem[Kriege et~al.(2016)Kriege, Giscard, and Wilson]{kriege2016on}
Nils~M. Kriege, Pierre{-}Louis Giscard, and Richard~C. Wilson.
\newblock On valid optimal assignment kernels and applications to graph
  classification.
\newblock In \emph{NeurIPS}, pages 1615--1623, 2016.

\bibitem[Kuramochi and Karypis(2004)]{kuramochi2004grew}
Michihiro Kuramochi and George Karypis.
\newblock {GREW-A} scalable frequent subgraph discovery algorithm.
\newblock In \emph{ICDM}, pages 439--442, 2004.

\bibitem[Lei et~al.(2017)Lei, Jin, Barzilay, and Jaakkola]{lei2017deriving}
Tao Lei, Wengong Jin, Regina Barzilay, and Tommi~S. Jaakkola.
\newblock Deriving neural architectures from sequence and graph kernels.
\newblock In \emph{{ICML}}, volume~70, pages 2024--2033, 2017.

\bibitem[Liu and Song(2022)]{liu2021graph}
Xin Liu and Yangqiu Song.
\newblock Graph convolutional networks with dual message passing for subgraph
  isomorphism counting and matching.
\newblock In \emph{AAAI}, pages 7594--7602, 2022.

\bibitem[Liu et~al.(2020)Liu, Pan, He, Song, Jiang, and Shang]{liu2020neural}
Xin Liu, Haojie Pan, Mutian He, Yangqiu Song, Xin Jiang, and Lifeng Shang.
\newblock Neural subgraph isomorphism counting.
\newblock In \emph{KDD}, pages 1959--1969, 2020.

\bibitem[Milo et~al.(2002)Milo, Shen-Orr, Itzkovitz, Kashtan, Chklovskii, and
  Alon]{milo2002network}
Ron Milo, Shai Shen-Orr, Shalev Itzkovitz, Nadav Kashtan, Dmitri Chklovskii,
  and Uri Alon.
\newblock Network motifs: simple building blocks of complex networks.
\newblock \emph{Science}, 298\penalty0 (5594):\penalty0 824--827, 2002.

\bibitem[Morris et~al.(2020)Morris, Rattan, and Mutzel]{morris2020weisfeiler}
Christopher Morris, Gaurav Rattan, and Petra Mutzel.
\newblock Weisfeiler and leman go sparse: Towards scalable higher-order graph
  embeddings.
\newblock In \emph{NeurIPS}, 2020.

\bibitem[Ren and Wang(2015)]{ren2015exploiting}
Xuguang Ren and Junhu Wang.
\newblock Exploiting vertex relationships in speeding up subgraph isomorphism
  over large graphs.
\newblock \emph{PVLDB}, 8\penalty0 (5):\penalty0 617--628, 2015.

\bibitem[Ribeiro and Silva(2010)]{ribeiro2010efficient}
Pedro Manuel~Pinto Ribeiro and Fernando M.~A. Silva.
\newblock Efficient subgraph frequency estimation with g-tries.
\newblock In \emph{{WABI}}, volume 6293, pages 238--249, 2010.

\bibitem[Schlichtkrull et~al.(2018)Schlichtkrull, Kipf, Bloem, van~den Berg,
  Titov, and Welling]{schlichtkrull2018modeling}
Michael~Sejr Schlichtkrull, Thomas~N. Kipf, Peter Bloem, Rianne van~den Berg,
  Ivan Titov, and Max Welling.
\newblock Modeling relational data with graph convolutional networks.
\newblock In \emph{ESWC}, volume 10843, pages 593--607, 2018.

\bibitem[Schomburg et~al.(2004)Schomburg, Chang, Ebeling, Gremse, Heldt, Huhn,
  and Schomburg]{schomburg2004brenda}
Ida Schomburg, Antje Chang, Christian Ebeling, Marion Gremse, Christian Heldt,
  Gregor Huhn, and Dietmar Schomburg.
\newblock Brenda, the enzyme database: updates and major new developments.
\newblock \emph{Nucleic acids research}, 32\penalty0 (suppl\_1):\penalty0
  D431--D433, 2004.

\bibitem[Schulz et~al.(2022)Schulz, Horv{\'{a}}th, Welke, and
  Wrobel]{schulz2022a}
Till~Hendrik Schulz, Tam{\'{a}}s Horv{\'{a}}th, Pascal Welke, and Stefan
  Wrobel.
\newblock A generalized weisfeiler-lehman graph kernel.
\newblock \emph{Machine Learning}, 111\penalty0 (7):\penalty0 2601--2629, 2022.

\bibitem[Shen et~al.(2019)Shen, Geng, Qin, Guo, Tang, Duan, Long, and
  Jiang]{shen2019multi}
Tao Shen, Xiubo Geng, Tao Qin, Daya Guo, Duyu Tang, Nan Duan, Guodong Long, and
  Daxin Jiang.
\newblock Multi-task learning for conversational question answering over a
  large-scale knowledge base.
\newblock In \emph{{EMNLP-IJCNLP}}, pages 2442--2451, 2019.

\bibitem[Shervashidze et~al.(2009)Shervashidze, Vishwanathan, Petri, Mehlhorn,
  and Borgwardt]{shervashidze2009efficient}
Nino Shervashidze, S.~V.~N. Vishwanathan, Tobias Petri, Kurt Mehlhorn, and
  Karsten~M. Borgwardt.
\newblock Efficient graphlet kernels for large graph comparison.
\newblock In \emph{AISTATS}, volume~5, pages 488--495, 2009.

\bibitem[Shervashidze et~al.(2011)Shervashidze, Schweitzer, van Leeuwen,
  Mehlhorn, and Borgwardt]{shervashidze2011weisfeiler}
Nino Shervashidze, Pascal Schweitzer, Erik~Jan van Leeuwen, Kurt Mehlhorn, and
  Karsten~M. Borgwardt.
\newblock Weisfeiler-lehman graph kernels.
\newblock \emph{Journal of Machine Learning Research}, 12:\penalty0 2539--2561,
  2011.

\bibitem[Tepeli et~al.(2020)Tepeli, {\"U}nal, Akdemir, and
  Tastan]{tepeli2020pamogk}
Yasin~Ilkagan Tepeli, Ali~Burak {\"U}nal, Furkan~Mustafa Akdemir, and Oznur
  Tastan.
\newblock Pamogk: a pathway graph kernel-based multiomics approach for patient
  clustering.
\newblock \emph{Bioinformatics}, 36\penalty0 (21):\penalty0 5237--5246, 2020.

\bibitem[Ullmann(1976)]{ullmann1976an}
Julian~R. Ullmann.
\newblock An algorithm for subgraph isomorphism.
\newblock \emph{Journal of the {ACM}}, 23\penalty0 (1):\penalty0 31--42, 1976.

\bibitem[Vashishth et~al.(2020)Vashishth, Sanyal, Nitin, and
  Talukdar]{vashishth2020composition}
Shikhar Vashishth, Soumya Sanyal, Vikram Nitin, and Partha~P. Talukdar.
\newblock Composition-based multi-relational graph convolutional networks.
\newblock In \emph{ICLR}, 2020.

\bibitem[Weisfeiler and Leman(1968)]{weisfeiler1968reduction}
Boris Weisfeiler and Andrei Leman.
\newblock The reduction of a graph to canonical form and the algebra which
  appears therein.
\newblock \emph{NTI, Series}, 2\penalty0 (9):\penalty0 12--16, 1968.

\bibitem[Wernicke(2005)]{wernicke2005a}
Sebastian Wernicke.
\newblock A faster algorithm for detecting network motifs.
\newblock In \emph{{WABI}}, volume 3692, pages 165--177, 2005.

\bibitem[Xu et~al.(2019)Xu, Hu, Leskovec, and Jegelka]{xu2019how}
Keyulu Xu, Weihua Hu, Jure Leskovec, and Stefanie Jegelka.
\newblock How powerful are graph neural networks?
\newblock In \emph{ICLR}, 2019.

\bibitem[Yan et~al.(2004)Yan, Yu, and Han]{yan2004graph}
Xifeng Yan, Philip~S. Yu, and Jiawei Han.
\newblock Graph indexing: {A} frequent structure-based approach.
\newblock In \emph{SIGMOD}, pages 335--346, 2004.

\bibitem[Ying et~al.(2018)Ying, He, Chen, Eksombatchai, Hamilton, and
  Leskovec]{ying2018graph}
Rex Ying, Ruining He, Kaifeng Chen, Pong Eksombatchai, William~L. Hamilton, and
  Jure Leskovec.
\newblock Graph convolutional neural networks for web-scale recommender
  systems.
\newblock In \emph{SIGKDD}, pages 974--983, 2018.

\bibitem[Ying et~al.(2020)Ying, Lou, You, Wen, Canedo, and
  Leskovec]{ying2020neural}
Rex Ying, Zhaoyu Lou, Jiaxuan You, Chengtao Wen, Arquimedes Canedo, and Jure
  Leskovec.
\newblock Neural subgraph matching.
\newblock \emph{arXiv preprint arXiv:2007.03092}, 2020.

\bibitem[Yu et~al.(2023)Yu, Liu, Fang, and Zhang]{yu2023learning}
Xingtong Yu, Zemin Liu, Yuan Fang, and Xinming Zhang.
\newblock Learning to count isomorphisms with graph neural networks.
\newblock In \emph{{AAAI}}, 2023.

\bibitem[Zhang et~al.(2013)Zhang, Song, Liu, Bu, and Chen]{zhang2013fast}
Luming Zhang, Mingli Song, Xiao Liu, Jiajun Bu, and Chun Chen.
\newblock Fast multi-view segment graph kernel for object classification.
\newblock \emph{Signal Processing}, 93\penalty0 (6):\penalty0 1597--1607, 2013.

\bibitem[Zhao et~al.(2017)Zhao, Yao, Li, Song, and Lee]{zhao2017meta}
Huan Zhao, Quanming Yao, Jianda Li, Yangqiu Song, and Dik~Lun Lee.
\newblock Meta-graph based recommendation fusion over heterogeneous information
  networks.
\newblock In \emph{SIGKDD}, pages 635--644, 2017.

\bibitem[Zhao et~al.(2012)Zhao, Wang, Butt, Khan, Kumar, and
  Marathe]{zhao2012sahad}
Zhao Zhao, Guanying Wang, Ali~R Butt, Maleq Khan, VS~Anil Kumar, and Madhav~V
  Marathe.
\newblock Sahad: Subgraph analysis in massive networks using hadoop.
\newblock In \emph{IPDPS}, pages 390--401, 2012.

\end{thebibliography}


\clearpage
\appendix
\section{Experiment Settings}
\label{appedix:exp_setting}

\subsection{Benchmarking Datasets}
\label{appendix:dataset}

We collect the most popular datasets (listed in Table~\ref{tab:stat_sic}) for graph classification, as graph properties are often determined by substructures within the graph, such as motifs.
Among the datasets, \textit{IMDB-BINARY} and \textit{IMDB-MULTI} consist of ego-networks derived from actor collaborations in the IMDB dataset. \textit{ENZYMES} represents macromolecules and their interactions in the bioinformatics field, which were collected from the BRENDA database~\cite{schomburg2004brenda}.
\textit{NCI109} consists of molecular compounds for interaction prediction. Additionally, we selected two synthetic homogeneous datasets, \textit{Erd\H{o}s-Renyi} and \textit{Regular}~\cite{chen2020can}, for benchmarking purposes.

\begin{table*}[!t]
    \tiny
    \centering
    \setlength\tabcolsep{3pt}
    \caption{Dataset statistics on subgraph isomorphism experiments. $\mathcal{P}$ and $\mathcal{G}$ corresponds to patterns and graphs.}
    \label{tab:stat_sic}
    \begin{tabular}{c|ccc|ccc|ccc|ccc|ccc|ccc}
    \toprule
        & \multicolumn{3}{c|}{\textit{Erd\H{o}s-Renyi}} & \multicolumn{3}{c|}{\textit{Regular}} & \multicolumn{3}{c|}{IMDB-BINARY} & \multicolumn{3}{c|}{IMDB-MULTI} & \multicolumn{3}{c|}{\textit{ENZYMES}} & \multicolumn{3}{c}{\textit{NCI109}} \\
        \toprule
        \# Train & \multicolumn{3}{c|}{6,000} & \multicolumn{3}{c|}{6,000} & \multicolumn{3}{c|}{1,332} & \multicolumn{3}{c|}{2,000} & \multicolumn{3}{c|}{8,400} & \multicolumn{3}{c}{6,875} \\
        \# Valid & \multicolumn{3}{c|}{4,000} & \multicolumn{3}{c|}{4,000} & \multicolumn{3}{c|}{1,336} & \multicolumn{3}{c|}{2,000} & \multicolumn{3}{c|}{8,400} & \multicolumn{3}{c}{6,880} \\
        \# Test & \multicolumn{3}{c|}{10,000} & \multicolumn{3}{c|}{10,000} & \multicolumn{3}{c|}{1,332} & \multicolumn{3}{c|}{2,000} & \multicolumn{3}{c|}{8,400} & \multicolumn{3}{c}{6,880} \\
        Avg. Subgraph Isomorphisms & \multicolumn{3}{c|}{51.655} & \multicolumn{3}{c|}{121.647} & \multicolumn{3}{c|}{37041.171} & \multicolumn{3}{c|}{33063.770} & \multicolumn{3}{c|}{25.110} & \multicolumn{3}{c}{7.703} \\
        \midrule
        & Max & \multicolumn{2}{c|}{Avg.} & Max & \multicolumn{2}{c|}{Avg.} & Max & \multicolumn{2}{c|}{Avg.} & Max & \multicolumn{2}{c|}{Avg.} & Max & \multicolumn{2}{c|}{Avg.} & Max & \multicolumn{2}{c}{Avg.} \\
        $|\mathcal{V}_{\mathcal{P}}|$ & 4 & \multicolumn{2}{c|}{3.8$\pm$0.4} & 4 & \multicolumn{2}{c|}{3.8$\pm$0.4} & 4 & \multicolumn{2}{c|}{3.75$\pm$0.4} & 4 & \multicolumn{2}{c|}{3.75$\pm$0.4} & 4 & \multicolumn{2}{c|}{3.9$\pm$0.3} & 4 & \multicolumn{2}{c}{4$\pm$0} \\
        $|\mathcal{E}_{\mathcal{P}}|$ & 10 & \multicolumn{2}{c|}{7.5$\pm$1.7} & 10 & \multicolumn{2}{c|}{7.5$\pm$1.7} & 10 & \multicolumn{2}{c|}{7.5$\pm$1.7} & 10 & \multicolumn{2}{c|}{7.5$\pm$1.7} & 10 & \multicolumn{2}{c|}{7.6$\pm$1.5} & 6 & \multicolumn{2}{c}{6$\pm$0} \\
        $|\mathcal{X}_{\mathcal{P}}|$ & 1 & \multicolumn{2}{c|}{1$\pm$0} & 1 & \multicolumn{2}{c|}{1$\pm$0} & 1 & \multicolumn{2}{c|}{1$\pm$0} & 1 & \multicolumn{2}{c|}{1$\pm$0} & 3 & \multicolumn{2}{c|}{2.1$\pm$0.5} & 3 & \multicolumn{2}{c}{3$\pm$0} \\
        $|\mathcal{Y}_{\mathcal{P}}|$ & 1 & \multicolumn{2}{c|}{1$\pm$0} & 1 & \multicolumn{2}{c|}{1$\pm$0} & 1 & \multicolumn{2}{c|}{1$\pm$0} & 1 & \multicolumn{2}{c|}{1$\pm$0} & 1 & \multicolumn{2}{c|}{1$\pm$0} & 1 & \multicolumn{2}{c}{1$\pm$0} \\
        $|\mathcal{V}_{\mathcal{G}}|$ & 10 & \multicolumn{2}{c|}{10$\pm$0} & 30 & \multicolumn{2}{c|}{18.8$\pm$7.4} & 136 & \multicolumn{2}{c|}{19.8$\pm$10.1} & 89 & \multicolumn{2}{c|}{13.0$\pm$8.5} & 126 & \multicolumn{2}{c|}{32.6$\pm$15.3} & 111 & \multicolumn{2}{c}{29.9$\pm$13.6} \\
        $|\mathcal{E}_{\mathcal{G}}|$ & 48 & \multicolumn{2}{c|}{27.0$\pm$6.1} & 90 & \multicolumn{2}{c|}{62.7$\pm$17.9} & 4,996 & \multicolumn{2}{c|}{386.1$\pm$442.4} & 5868 & \multicolumn{2}{c|}{263.7$\pm$443.1} & 298 & \multicolumn{2}{c|}{124.3$\pm$51.0} & 238 & \multicolumn{2}{c}{64.6$\pm$29.9} \\
        $|\mathcal{X}_{\mathcal{G}}|$ & 1 & \multicolumn{2}{c|}{1$\pm$0} & 1 & \multicolumn{2}{c|}{1$\pm$0} & 1 & \multicolumn{2}{c|}{1$\pm$0} & 3 & \multicolumn{2}{c|}{2.1$\pm$0.3} & 1 & \multicolumn{2}{c|}{1$\pm$0} & 38 & \multicolumn{2}{c}{5.2$\pm$4.0} \\
        $|\mathcal{Y}_{\mathcal{G}}|$ & 1 & \multicolumn{2}{c|}{1$\pm$0} & 1 & \multicolumn{2}{c|}{1$\pm$0} & 1 & \multicolumn{2}{c|}{1$\pm$0} & 1 & \multicolumn{2}{c|}{1$\pm$0} & 1 & \multicolumn{2}{c|}{1$\pm$0} & 1 & \multicolumn{2}{c}{1$\pm$0} \\
        \bottomrule
    \end{tabular}
\end{table*}

In order to obtain meaningful and challenging predictions, we enumerate all permutations of vertex labels and permutations of edge labels from the 3-stars, triangles, tailed triangles, and chordal cycles.
Furthermore, to improve the quality of the data, we have filtered out patterns with an average frequency of less than 1.0 across the entire dataset.
The program can be summarized by Algorithm~\ref{alg:data_construct}. We consistently enumerate the permutations of vertex labels and edge labels, and we check the validity of heterogeneous patterns.
The subgraph isomorphism counting is efficiently performed by the VF2 algorithm~\cite{cordella2004a} in parallel.
The construction time for each dataset is less than three hours.

\begin{algorithm}[!t]
    \caption{Benchmarking data construction.}
    \label{alg:data_construct}
    {\footnotesize
        \begin{algorithmic}[1]
        \INPUT a set of directed graphs $\mathcal{G}_{1}, \mathcal{G}_{2}, \cdots, \mathcal{G}_{D}$, a homogeneous pattern structure $\mathcal{P}'$
        \STATE construct a set of directed homogeneous graphs $\mathcal{G}'_{1}, \mathcal{G}'_{2}, \cdots, \mathcal{G}'_{D}$ where vertex labels and edge labels are dropped
        \STATE conduct subgraph isomorphism counting for pattern $\mathcal{P}'$ over homogeneous graphs $\mathcal{G}'_{1}, \mathcal{G}'_{2}, \cdots, \mathcal{G}'_{D}$
        \STATE create a heterogeneous pattern set $\mathcal{S}$
        \IF{the average counting per graph is less than or equal to 1.0}
        \STATE get the average vertex label integer $\lceil \overline{x} \rceil$ and average edge label integer $\lceil \overline{y} \rceil$
        \FOR{iter vertex label assignment $\mathcal{X}_{\mathcal{P}}$ from $\{1, 2, \cdots, \lceil \overline{x} \rceil \}^{|\mathcal{V}_{\mathcal{P}'}|}$}
            \FOR{iter edge label assignment $\mathcal{Y}_{\mathcal{P}}$ from $\{1, 2, \cdots, \lceil \overline{y} \rceil \}^{|\mathcal{E}_{\mathcal{P}'}|}$}
                \STATE construct a heterogeneous pattern $\mathcal{P}$ with the same structure of $\mathcal{P}'$, and two label assignments $\mathcal{X}_{\mathcal{P}}$ and $\mathcal{Y}_{\mathcal{P}}$
                \STATE conduct subgraph isomorphism counting for pattern $\mathcal{P}$ over graphs $\mathcal{G}_{1}, \mathcal{G}_{2}, \cdots, \mathcal{G}_{D}$
                \IF{the average counting per graph is greater than 1.0}
                    \STATE add current heterogeneous pattern 
 $\mathcal{P}$ to the pattern set $\mathcal{S}$
                \ENDIF
            \ENDFOR
        \ENDFOR
        \ENDIF
        \OUTPUT the heterogeneous pattern set $\mathcal{S}$
        \end{algorithmic}
    }
\end{algorithm}

\subsection{Parameter Selection}
\label{appendix:parameter}

Graph kernels do not require any parameters to be tuned. 
However, the polynomial kernel and the Gaussian kernel both have hyper-parameters
Specifically, for polynomial kernels, we fix the power of the polynomial to 3 and tune the factor of the radix among $\{2e\text{-}5, 2e\text{-}4, \cdots, 1\}$;
for Gaussian kernels, we search for the hyper-parameter $2\sigma^2$ in the range of $\{1e\text{-}5, 1e\text{-}4, \cdots, 1e5\}$.
When using SVM, we tune the regularization parameter $C$ in the range of $\{1e\text{-}2, \cdots, 1e4\}$; when using Ridge, we tune the regularization parameter $\alpha$ in the range of $\{1e\text{-}4, 1e\text{-}3, \cdots, 1e2\}$.
Models are trained and selected based on the validation set, with the mean squared error (MSE) serving as the evaluation metric.
The seed is fixed as 2023, and we do not observe the performance change with different seeds.

\section{Experimental Results}
\label{appendix:exp_result}

\subsection{Trivial Baselines and Neural Networks}

Two trivial baselines ignore the input data but always make predictions based on the training statistics.
For example, the \textbf{Zero} baseline always returns zeros because a random graph is highly unlikely to be matched by a heterogeneous pattern.
The \textbf{Avg} baseline tends to predict the average count based on the training data, as the maximum expectation.
As shown in Table~\ref{tab:neural}, we can observe that predicting zeros usually yields better absolute errors than predicting the average, indicating the difficulty of the isomorphism counting task.
Furthermore, the errors can reach values in the hundreds for both synthetic and real-life data, highlighting the challenge involved.

As a problem of learning to predict, we compare our graph kernel methods with neural networks.
\citet{liu2020neural} have released an implementation of a neural subgraph isomorphism counting framework\footnote{https://github.com/HKUST-KnowComp/NeuralSubgraphCounting}, which we directly employ to report the results of \textbf{CNN}~\cite{kim2014convolutional}, \textbf{LSTM}~\cite{hochreiter1997long}, \textbf{TXL}~\cite{transformerxl2019dai}, \textbf{RGCN}~\cite{schlichtkrull2018modeling}, \textbf{RGIN}~\cite{liu2020neural}, and \textbf{CompGCN}~\cite{vashishth2020composition} in Table~\ref{tab:neural}.
Graph neural networks (RGCN, RGIN, and CompGCN) outperform sequence models.
However, sequence models can still provide relatively accurate predictions when the data is extremely challenging, such as in the case of \textit{IMDB-BINARY} and \textit{IMDB-MULTI}.
This encourages researchers to generalize graph neural networks to handle complicated cases, especially considering that RGIN consistently achieves the best results on other datasets.

\begin{table*}[!t]
    \scriptsize 
    \centering 
    \setlength\tabcolsep{2.3pt} 
    \caption{Performance on subgraph isomorphism counting with naive baselines and neural networks.}
    \label{tab:neural}
    \begin{tabular}{l|cc|cc|cc|cc|cc|cc}
    \toprule
        \multicolumn{1}{c|}{\multirow{3}{*}{Models}} & \multicolumn{8}{c|}{Homogeneous}  & \multicolumn{4}{c}{Heterogeneous} \\
        \cline{2-13}
        & \multicolumn{2}{c|}{\textit{Erd\H{o}s-Renyi}} & \multicolumn{2}{c|}{\textit{Regular}} & \multicolumn{2}{c|}{IMDB-BINARY} & \multicolumn{2}{c|}{IMDB-MULTI} & \multicolumn{2}{c|}{\textit{ENZYMES}} & \multicolumn{2}{c}{\textit{NCI109}} \\
        & RMSE & MAE & RMSE & MAE & RMSE & MAE & RMSE & MAE & RMSE & MAE & RMSE & MAE \\
        \midrule
        \multirow{1}{*}{Zero} & 92.532 & 51.655 & 198.218 & 121.647 & 138262.003 & 37041.171 & 185665.874 & 33063.770 & 64.658 & 25.110 & 16.882 & 7.703 \\
        \multirow{1}{*}{Avg}  & 121.388 & 131.007  & 156.515 & 127.211 & 133228.554 & 54178.671 & 182717.385 & 53398.301 & 59.589 & 31.577 & 14.997 & 8.622 \\
        \hline
        \multirow{1}{*}{CNN} & 20.386 & 13.316 & 37.192 & 27.268 & 4808.156 & 1570.293 & 4185.090 & 1523.731 & 16.752 & 7.720 & 3.096 & 1.504 \\
        \multirow{1}{*}{LSTM} & 14.561 & 9.949 & 14.169 & 10.064 & 10596.339 & 2418.997 & 10755.528 & 1925.363 & 20.211 & 8.841 & 4.467 & 2.234 \\
        \multirow{1}{*}{TXL} & 10.861 & 7.105 & 15.263 & 10.721 & 15369.186 & 3170.290 & 19706.248 & 3737.862 & 25.912 & 11.284 & 5.482 & 2.823 \\
        \multirow{1}{*}{RGCN} & 9.386 & 5.829 & 14.789 & 9.772 & 46074.355 & 13498.414 & 69485.242 & 12137.598 & 23.715 & 11.407 & 1.217 & 0.622 \\
        \multirow{1}{*}{RGIN} & 6.063 & 3.712 & 13.554 & 8.580 & 31058.764 & 6445.103 & 26546.882 & 4508.339 & 8.119 & 3.783 & 0.511 & 0.292 \\
        \multirow{1}{*}{CompGCN} & 6.706 & 4.274 & 14.174 & 9.685 & 32145.713 & 8576.071 & 26523.880 & 7745.702 & 14.985 & 6.438 & 1.271 & 0.587 \\
        \bottomrule
    \end{tabular}
\end{table*}

\subsection{SVM vs. Ridge}
\label{appendix:svm_ridge}

\begin{figure*}[!h]
    \centering
    \begin{subfigure}[b]{0.48\linewidth}
        \centering
        \includegraphics[height=2.0in]{figures/svm_linear.pdf}
        \vspace{-0.1in}
        \caption{SVM, Linear.}
    \end{subfigure}
    \begin{subfigure}[b]{0.48\linewidth}
        \centering
        \includegraphics[height=2.0in]{figures/ridge_linear.pdf}
        \vspace{-0.1in}
        \caption{Ridge, Linear.}
    \end{subfigure}
    \begin{subfigure}[b]{0.48\linewidth}
        \centering
        \includegraphics[height=2.0in]{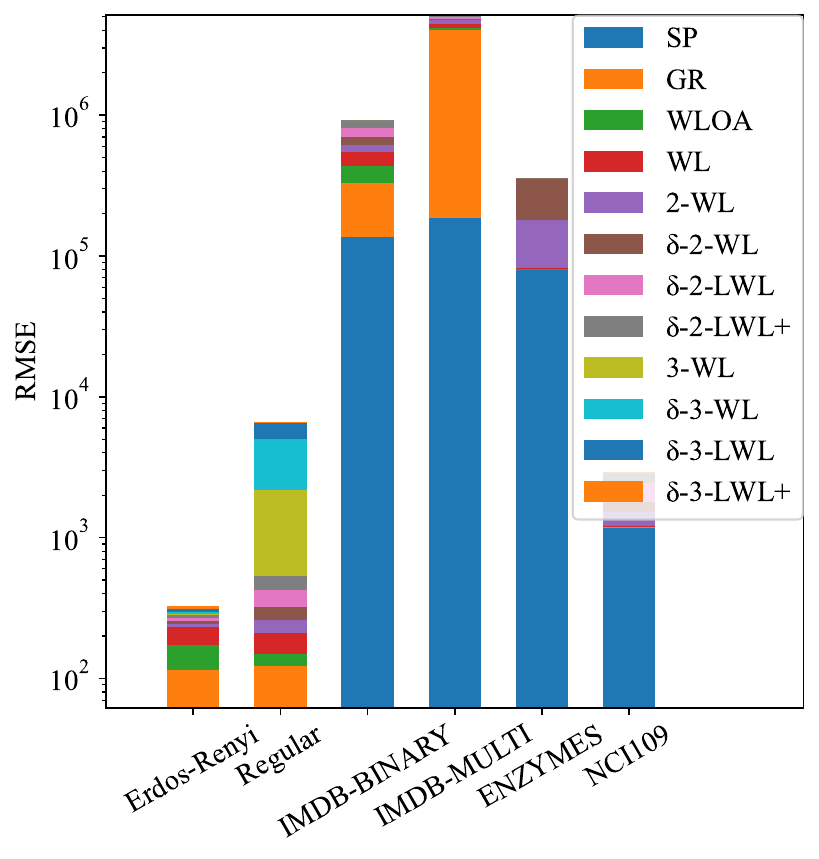}
        \vspace{-0.1in}
        \caption{SVM, Poly.}
    \end{subfigure}
    \begin{subfigure}[b]{0.48\linewidth}
        \centering
        \includegraphics[height=2.0in]{figures/ridge_poly.pdf}
        \vspace{-0.1in}
        \caption{Ridge, Poly.}
    \end{subfigure}
    \begin{subfigure}[b]{0.48\linewidth}
        \centering
        \includegraphics[height=2.0in]{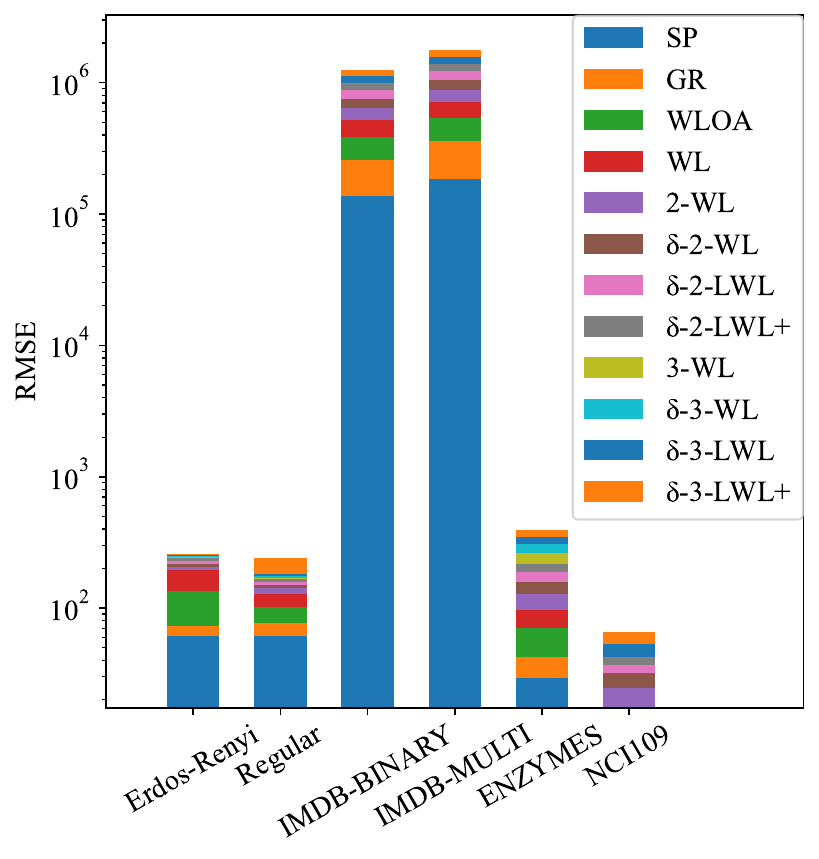}
        \vspace{-0.1in}
        \caption{SVM, RBF.}
    \end{subfigure}
    \begin{subfigure}[b]{0.48\linewidth}
        \centering
        \includegraphics[height=2.0in]{figures/ridge_rbf.pdf}
        \vspace{-0.1in}
        \caption{Ridge, RBF.}
    \end{subfigure}
    \caption{Performance on subgraph isomorphism counting with SVM and Ridge with kernel tricks, where ``Poly'' and ``RBF'' indicatet the polynomial and Gaussian kernels, and the out-of-memory (``OOM'') and not-a-number (``NaN'') are regarded as zeros in the plots.}
    \label{fig:appendix_svm_ridge}
\end{figure*}

We compare the performance of SVM and Ridge (precisely, Kernel Ridge) regression using kernel tricks for subgraph isomorphism counting, which is a common practice to evaluate the performance of these two regressors.
The results are shown in Figure~\ref{fig:appendix_svm_ridge}, and it can be observed that the performance of the two regressors is comparable, with Ridge performing slightly better.
There are two main reasons for this.
Firstly, Ridge is solved using Cholesky factorization in the closed form, which typically achieves better convergence than the iterative optimization used in SVM.
Secondly, the objective of Ridge is to minimize the sum of squared errors, which is more straightforward and suitable for the regression task compared to the $\epsilon$-insensitive loss function used in SVM.
Given the superior performance of Ridge, we report only the results of Ridge in the following experiments to save space.

In addition to the comparison between SVM and Ridge, we also observed an increase in errors with the polynomial kernel trick.
While the number of matched subgraphs is typically small, it can be very large for certain structures such as complete graphs or cliques.
The polynomial kernel trick can easily lead to fluctuations due to extreme values, resulting in performance fluctuations and even overflow.
Therefore, our focus primarily lies on the original kernels and the Gaussian kernel trick.

\subsection{Normalization}
\label{appendix:normalization}

Postprocessing is a common technique used in regression, such as feature normalization to reduce variance and regularization to prevent overfitting.
A common way to normalize the Gram matrix involves dividing each element by the square root of the product of the corresponding row and column norms, resulting in $\bm{K}_{\text{norm}_{i,j}} = \bm{K}_{i,j} / \sqrt{\bm{K}_{i,i} \cdot \bm{K}_{j,j}}$.
We investigate the impact of normalization on graph kernels, which are illustrated in Table~\ref{tab:ridge_norm}.
As expected, normalization significantly harms the performance, regardless of whether the Gaussian kernel is applied or not.
We analyze the reason behind this phenomenon using the example of the linear kernel.
The graph kernel matrix normalization is equivalent to the normalization of the graph representation.
Assume the obtained graph representaion is $\bm{g}_{i}$, then the corresponding normalized graph representation is $\bm{g}_{\text{norm}}$.
The element of $\bm{g}_{i}[j]$ is $k(\mathcal{G}_i, \mathcal{G}_j)$, which is calculated by the inner product of feature vectors.
The unnormalized feature vector is equivalent to the concatenation of the pattern histogram.
Thus, $k(\mathcal{G}_i, \mathcal{G}_j)$ is the inner product of the histograms of the two graphs, i.e., $\biggm\langle \biggm\Vert_{t=0}^{T} \bm C_{\mathcal{V}_{\mathcal{G}_i}}^{(t)} , \biggm\Vert_{t=0}^{T} \bm C_{\mathcal{V}_{\mathcal{G}_j}}^{(t)} \biggm\rangle$ for WL kernel.
The normalized Gram matrix actually calculates the cosine similarity, i.e., $k_{\text{norm}}(\mathcal{G}_i, \mathcal{G}_j) = \frac{\biggm\langle \biggm\Vert_{t=0}^{T} \bm C_{\mathcal{V}_{\mathcal{G}_i}}^{(t)} , \biggm\Vert_{t=0}^{T} \bm C_{\mathcal{V}_{\mathcal{G}_j}}^{(t)} \biggm\rangle} {\sqrt{\biggm\langle \biggm\Vert_{t=0}^{T} \bm C_{\mathcal{V}_{\mathcal{G}_i}}^{(t)} , \biggm\Vert_{t=0}^{T} \bm C_{\mathcal{V}_{\mathcal{G}_i}}^{(t)} \biggm\rangle \cdot \biggm\langle \biggm\Vert_{t=0}^{T} \bm C_{\mathcal{V}_{\mathcal{G}_j}}^{(t)} , \biggm\Vert_{t=0}^{T} \bm C_{\mathcal{V}_{\mathcal{G}_j}}^{(t)} \biggm\rangle}}$.
Therefore, the normalized Gram matrix measures the cosine similarity of two color distributions, rather than the similarity of two graph structures.
It is not surprising that the normalized Gram matrix performs worse than the unnormalized one, as the information inside distributions is less than the information inside histograms.

\begin{table*}[!t]
    \scriptsize 
    \centering 
    \setlength\tabcolsep{2.3pt} 
    \caption{Performance comparison with or without normalization.}
    \label{tab:ridge_norm}
    \begin{tabular}{l|cc|cc|cc|cc|cc|cc}
    \toprule
        \multicolumn{1}{c|}{\multirow{3}{*}{Models}} & \multicolumn{8}{c|}{Homogeneous}  & \multicolumn{4}{c}{Heterogeneous} \\
        \cline{2-13}
        & \multicolumn{2}{c|}{\textit{Erd\H{o}s-Renyi}} & \multicolumn{2}{c|}{\textit{Regular}} & \multicolumn{2}{c|}{IMDB-BINARY} & \multicolumn{2}{c|}{IMDB-MULTI} & \multicolumn{2}{c|}{\textit{ENZYMES}} & \multicolumn{2}{c}{\textit{NCI109}} \\
        & RMSE & MAE & RMSE & MAE & RMSE & MAE & RMSE & MAE & RMSE & MAE & RMSE & MAE \\
        \midrule
        \multicolumn{13}{c}{\textbf{Ridge, Linear}} \\
        \hline
        SP & 58.721 & 34.606 & 60.375 & 41.110 & 131672.705 & 56058.593 & 181794.702 & 54604.564 & 43.007 & 14.422 & 4.824 & 2.268 \\
        GR & 14.067 & 7.220 & 23.775 & 12.172 & \bf 30527.764 & 7894.568 & \bf 30980.135 & 6054.027 & 14.557 & 5.595 & 5.066 & 2.066 \\
        WLOA & 58.719 & 34.635 & 25.905 & 17.003 & 96887.226 & 28849.659 & 117828.698 & 25808.362 & 28.911 & 11.807 & 3.142 & 1.142 \\
        WL & 58.719 & 34.635 & 56.045 & 33.891 & 107500.276 & 41523.359 & 147822.358 & 49244.753 & 46.466 & 14.920 & \bf 1.896 & \bf 0.746 \\
        2-WL & 10.452 & 5.561 & 12.353 & 7.906 & 34734.939 & 9161.265 & 47075.541 & 13751.520 & 26.903 & 11.018 & 7.003 & 3.060 \\
        $\delta$-2-WL & 9.921 & 4.164 & 8.751 & 5.663 & 33336.019 & 9265.499 & 47075.541 & 13751.520 & 27.528 & 11.286 & 6.910 & 3.039 \\
        $\delta$-2-LWL & 11.342 & 4.757 & 11.020 & 7.230 & 38507.321 & 16105.742 & 47075.541 & 13751.520 & 54.915 & 10.079 & 2.605 & 1.072 \\
        $\delta$-2-LWL$^{+}$ & 11.132 & 4.687 & 11.795 & 7.703 & 38507.321 & 16105.742 & 47075.541 & 13751.520 & 89.581 & 10.911 & 2.584 & 1.068 \\
        3-WL & \bf 4.096 & \bf 1.833 & 4.038 & 2.330 & OOM & OOM & OOM & OOM & 335.940 & 13.790 & 9.721 & 3.314 \\
        $\delta$-3-WL & 4.214 & 1.840 & 4.092 & 2.361 & OOM & OOM & OOM & OOM & 387.816 & 15.573 & 9.712 & 3.290 \\
        $\delta$-3-LWL & 5.163 & 1.930 & \bf 3.975 & \bf 2.277 & 43894.672 & 8029.452 & 76218.966 & 9022.754 & 1727.556 & 42.346 & 3.872 & 1.375 \\
        $\delta$-3-LWL$^{+}$ & 5.151 & 1.931 & 60.375 & 41.110 & 39237.071 & 7240.730 & 76218.966 & 9022.754 & 1719.251 & 42.626 & 12.488 & 6.501 \\
        \hline
        \multicolumn{13}{c}{\textbf{Ridge, Linear w/ normalization}} \\
        \hline
        SP & 58.721 & 34.606 & 60.375 & 41.110 & 131672.705 & 56058.593 & 181794.702 & 54604.564 & 37.066 & 15.686 & 7.776 & 3.572 \\
        GR & 45.922 & 23.308 & 53.449 & 29.348 & 119288.171 & 47374.424 & 181431.698 & 54857.785 & 32.970 & 13.297 & 9.611 & 4.093 \\
        WLOA & 58.719 & 34.635 & 25.905 & 17.003 & 96120.928 & 28554.467 & 135879.827 & 31445.034 & 29.247 & 12.023 & 3.594 & 1.295 \\
        WL & 58.721 & 34.606 & 60.375 & 41.110 & 131672.705 & 56058.577 & 181794.702 & 54604.564 & 27.789 & 11.801 & 4.581 & 1.861 \\
        2-WL & 10.806 & 5.708 & 12.350 & 7.941 & 87428.241 & 29580.956 & 78615.331 & 16385.534 & 35.913 & 15.344 & 8.639 & 4.250 \\
        $\delta$-2-WL & 10.143 & 4.129 & 8.999 & 5.874 & 87894.665 & 31526.343 & 80061.204 & 17979.721 & 36.085 & 15.397 & 8.879 & 4.324 \\
        $\delta$-2-LWL & 11.934 & 5.024 & 15.585 & 8.976 & 81849.707 & 28816.597 & 82529.598 & 15951.174 & 27.863 & 11.174 & 5.121 & 2.033 \\
        $\delta$-2-LWL$^{+}$ & 11.701 & 4.942 & 15.440 & 8.967 & 81280.195 & 28569.835 & 83319.396 & 17110.823 & 27.868 & 11.183 & 5.128 & 2.037 \\
        3-WL & 4.642 & 2.155 & 11.773 & 7.073 & OOM & OOM & OOM & OOM & 32.908 & 14.018 & OOM & OOM \\
        $\delta$-3-WL & 4.745 & 2.207 & 12.000 & 7.147 & OOM & OOM & OOM & OOM & 32.851 & 13.989 & OOM & OOM \\
        $\delta$-3-LWL & 4.918 & 2.308 & 14.073 & 8.579 & 80474.058 & 25664.344 & 72950.070 & 14878.074 & 31.425 & 12.998 & 3.872 & 1.613 \\
        $\delta$-3-LWL$^{+}$ & 4.917 & 2.306 & 60.375 & 41.110 & 81622.690 & 26328.283 & 72715.240 & 14903.663 & 31.416 & 13.009 & 12.488 & 6.501 \\
        \hline
        \multicolumn{13}{c}{\textbf{Ridge, RBF}} \\
        \hline
        SP & 58.721 & 34.606 & 60.375 & 41.110 & 131672.705 & 56058.593 & 181794.702 & 54604.564 & 38.945 & 14.712 & 5.474 & 2.224 \\
        GR & 11.670 & 5.663 & 12.488 & 5.012 & 42387.021 & \bf 5110.985 & 41171.761 & \bf 4831.495 & \bf12.883 & \bf5.073 & 4.804 & 1.944 \\
        WLOA & 58.719 & 34.635 & 25.906 & 17.002 & 92733.105 & 28242.033 & 137300.092 & 34067.513 & 32.827 & 12.230 & 3.215 & 1.261 \\
        WL & 58.719 & 34.635 & 25.905 & 17.003 & 109418.159 & 32350.523 & 112515.690 & 25035.268 & 26.313 & 10.933 & 2.227 & 0.837 \\
        2-WL & 11.010 & 5.926 & 12.618 & 8.317 & 40412.745 & 5351.789 & 21910.109 & 2982.532 & 32.424 & 12.948 & 7.164 & 3.271 \\
        $\delta$-2-WL & 10.500 & 4.630 & 9.316 & 6.207 & 40412.745 & 5351.789 & 21910.109 & 2982.532 & 32.518 & 13.045 & 7.409 & 3.287 \\
        $\delta$-2-LWL & 11.788 & 5.004 & 8.643 & 5.730 & 40412.745 & 5351.789 & 21910.109 & 2982.532 & 29.560 & 11.878 & 5.010 & 1.806 \\
        $\delta$-2-LWL$^{+}$ & 11.659 & 4.936 & 8.495 & 5.634 & 40412.745 & 5351.789 & 21910.109 & 2982.532 & 30.525 & 11.977 & 5.001 & 1.799 \\
        3-WL & 4.949 & 2.568 & 4.631 & 2.783 & OOM & OOM & OOM & OOM & 43.909 & 18.509 & OOM & OOM \\
        $\delta$-3-WL & 4.896 & 2.536 & 4.567 & 2.745 & OOM & OOM & OOM & OOM & 43.908 & 18.509 & OOM & OOM \\
        $\delta$-3-LWL & 16.720 & 2.980 & 5.356 & 3.149 & 89532.736 & 21918.757 & 91445.323 & 17703.656 & 43.909 & 18.510 & 10.925 & 5.320 \\
        $\delta$-3-LWL$^{+}$ & 16.721 & 2.972 & 60.375 & 41.110 & 89532.736 & 21918.757 & 91445.323 & 17703.656 & 43.908 & 18.513 & 12.488 & 6.501 \\
        \hline
        \multicolumn{13}{c}{\textbf{Ridge, RBF w/ normalization}} \\
        \hline
        SP & 58.721 & 34.606 & 60.375 & 41.110 & 131672.705 & 56058.593 & 181794.702 & 54604.564 & 30.476 & 12.663 & 6.093 & 2.703 \\
        GR & 48.525 & 20.658 & 41.767 & 19.466 & 112530.277 & 44929.590 & 119919.313 & 28394.513 & 32.496 & 10.665 & 8.561 & 3.470 \\
        WLOA & 58.719 & 34.635 & 25.905 & 17.003 & 93888.723 & 28559.126 & 135642.945 & 31846.785 & 28.347 & 11.636 & 3.682 & 1.322 \\
        WL & 58.721 & 34.606 & 60.375 & 41.110 & 131112.523 & 54003.979 & 181794.702 & 54604.564 & 27.074 & 11.366 & 4.344 & 1.699 \\
        2-WL & 11.019 & 5.796 & 12.467 & 8.064 & 74732.522 & 20986.727 & 72753.757 & 18058.778 & 29.779 & 12.300 & 7.652 & 3.580 \\
        $\delta$-2-WL & 10.382 & 4.248 & 9.203 & 5.988 & 74732.522 & 20986.727 & 72753.757 & 18058.778 & 30.062 & 12.331 & 7.512 & 3.555 \\
        $\delta$-2-LWL & 12.112 & 5.137 & 11.271 & 7.170 & 74732.522 & 20986.727 & 79586.593 & 16979.269 & 26.008 & 10.462 & 3.926 & 1.740 \\
        $\delta$-2-LWL$^{+}$ & 11.871 & 5.054 & 11.220 & 7.132 & 74732.522 & 20986.727 & 76561.843 & 17602.345 & 25.997 & 10.454 & 3.921 & 1.744 \\
        3-WL & 4.686 & 2.175 & 6.113 & 3.315 & OOM & OOM & OOM & OOM & 30.498 & 12.341 & OOM & OOM \\
        $\delta$-3-WL & 4.731 & 2.281 & 5.611 & 3.137 & OOM & OOM & OOM & OOM & 30.886 & 12.448 & OOM & OOM \\
        $\delta$-3-LWL & 4.685 & 2.232 & 5.780 & 3.228 & 79781.673 & 26393.124 & 73428.449 & 15311.455 & 28.726 & 11.354 & 3.288 & 1.417 \\
        $\delta$-3-LWL$^{+}$ & 4.692 & 2.232 & 60.375 & 41.110 & 79334.836 & 26283.169 & 73217.154 & 15336.018 & 28.728 & 11.352 & 12.488 & 6.501 \\
        \bottomrule
    \end{tabular}
\end{table*}

\subsection{Neighborhood Information Extraction}
\label{appendix:nie}

Explicit neighborhood information extraction (NIE) is a crucial component for handling homogeneous data by providing edge colors.
As demonstrated in Table~\ref{tab:ridge_nie}, incorporating NIE consistently enhances the performance of both linear and RBF kernels.
Specifically, the RBF kernel combined with NIE proves to be more effective for homogeneous data, while the linear kernel shows substantial improvement when applied to heterogeneous data.
The most significant enhancements are observed on the highly challenging \textit{IMDB-BINARY} and \textit{IMDB-MULTI} datasets, where the RMSE is significantly reduced from 30,527.764 to 757.736 and from 21,910.109 to 833.037, respectively.
However, for the remaining two heterogeneous datasets, neighborhood information does not outperform as well as the GR kernel in the \textit{ENZYMES} dataset.
Upon analyzing the differences between the GR kernel and our proposed NIE, we find that high-order topologies such as triangles and wedges may be more powerful than first-order topologies.
However, it is noteworthy that the 3-WL-family kernels may perform poorly on heterogeneous data.
These findings serve as a foundation for further research and advancements in the field of graph kernels.

\begin{table*}[!t]
    \scriptsize 
    \centering 
    \setlength\tabcolsep{2.3pt} 
    \caption{Performance comparison with or without neighborhood information extraction.}
    \label{tab:ridge_nie}
    \begin{tabular}{l|cc|cc|cc|cc|cc|cc}
    \toprule
        \multicolumn{1}{c|}{\multirow{3}{*}{Models}} & \multicolumn{8}{c|}{Homogeneous}  & \multicolumn{4}{c}{Heterogeneous} \\
        \cline{2-13}
        & \multicolumn{2}{c|}{\textit{Erd\H{o}s-Renyi}} & \multicolumn{2}{c|}{\textit{Regular}} & \multicolumn{2}{c|}{IMDB-BINARY} & \multicolumn{2}{c|}{IMDB-MULTI} & \multicolumn{2}{c|}{\textit{ENZYMES}} & \multicolumn{2}{c}{\textit{NCI109}} \\
        & RMSE & MAE & RMSE & MAE & RMSE & MAE & RMSE & MAE & RMSE & MAE & RMSE & MAE \\
        \midrule
        \multicolumn{13}{c}{\textbf{Ridge, Linear}} \\
        \hline 
        SP & 58.721 & 34.606 & 60.375 & 41.110 & 131672.705 & 56058.593 & 181794.702 & 54604.564 & 43.007 & 14.422 & 4.824 & 2.268 \\
        GR & 14.067 & 7.220 & 23.775 & 12.172 & 30527.764 & 7894.568 & 30980.135 & 6054.027 & 14.557 & 5.595 & 5.066 & 2.066 \\
        WLOA & 58.719 & 34.635 & 25.905 & 17.003 & 96887.226 & 28849.659 & 117828.698 & 25808.362 & 28.911 & 11.807 & 3.142 & 1.142 \\
        WL & 58.719 & 34.635 & 56.045 & 33.891 & 107500.276 & 41523.359 & 147822.358 & 49244.753 & 46.466 & 14.920 & 1.896 & 0.746 \\
        2-WL & 10.452 & 5.561 & 12.353 & 7.906 & 34734.939 & 9161.265 & 47075.541 & 13751.520 & 26.903 & 11.018 & 7.003 & 3.060 \\
        $\delta$-2-WL & 9.921 & 4.164 & 8.751 & 5.663 & 33336.019 & 9265.499 & 47075.541 & 13751.520 & 27.528 & 11.286 & 6.910 & 3.039 \\
        $\delta$-2-LWL & 11.342 & 4.757 & 11.020 & 7.230 & 38507.321 & 16105.742 & 47075.541 & 13751.520 & 54.915 & 10.079 & 2.605 & 1.072 \\
        $\delta$-2-LWL$^{+}$ & 11.132 & 4.687 & 11.795 & 7.703 & 38507.321 & 16105.742 & 47075.541 & 13751.520 & 89.581 & 10.911 & 2.584 & 1.068 \\
        3-WL & \bf 4.096 & \bf 1.833 & 4.038 & 2.330 & OOM & OOM & OOM & OOM & 335.940 & 13.790 & 9.721 & 3.314 \\
        $\delta$-3-WL & 4.214 & 1.840 & 4.092 & 2.361 & OOM & OOM & OOM & OOM & 387.816 & 15.573 & 9.712 & 3.290 \\
        $\delta$-3-LWL & 5.163 & 1.930 & \bf 3.975 & \bf 2.277 & 43894.672 & 8029.452 & 76218.966 & 9022.754 & 1727.556 & 42.346 & 3.872 & 1.375 \\
        $\delta$-3-LWL$^{+}$ & 5.151 & 1.931 & 60.375 & 41.110 & 39237.071 & 7240.730 & 76218.966 & 9022.754 & 1719.251 & 42.626 & 12.488 & 6.501 \\
        \hline 
        \multicolumn{13}{c}{\textbf{Ridge, Linear w/ NIE}} \\
        \hline 
        SP & 58.721 & 34.606 & 60.375 & 41.110 & 131672.705 & 56058.593 & 181794.702 & 54604.564 & 43.007 & 14.422 & 4.824 & 2.268 \\
        GR & 14.067 & 7.220 & 23.775 & 12.172 & 30527.764 & 7894.568 & 30980.135 & 6054.027 & 14.557 & 5.595 & 5.066 & 2.066 \\
        WLOA & 58.719 & 34.635 & 25.905 & 17.003 & 33625.086 & 6009.372 & 20858.288 & 2822.391 & 23.478 & 10.037 & 3.203 & 1.133 \\
        WL & 58.719 & 34.635 & 56.045 & 33.891 & 66414.032 & 17502.328 & 70013.402 & 13266.318 & 20.971 & 8.672 & 1.772 & 0.704 \\
        2-WL & 10.452 & 5.561 & 12.353 & 7.906 & 34135.093 & 6275.320 & 47069.352 & 13669.964 & 211.105 & 13.200 & 8.747 & 3.051 \\
        $\delta$-2-WL & 9.921 & 4.164 & 8.751 & 5.663 & 14914.025 & 3671.681 & 47069.434 & 13671.226 & 238.306 & 14.007 & 7.369 & 2.954 \\
        $\delta$-2-LWL & 11.342 & 4.757 & 11.020 & 7.230 & 26549.602 & 4997.981 & 39932.609 & 10177.426 & 243.690 & 9.925 & \bf 1.259 & \bf 0.539 \\
        $\delta$-2-LWL$^{+}$ & 11.132 & 4.687 & 11.795 & 7.703 & 28183.800 & 5240.118 & 37676.903 & 9930.398 & 97.024 & 7.191 & 1.266 & 0.545 \\
        3-WL & \bf 4.096 & \bf 1.833 & 4.038 & 2.330 & OOM & OOM & OOM & OOM & OOM & OOM & OOM & OOM \\
        $\delta$-3-WL & 4.214 & 1.840 & 4.092 & 2.361 & OOM & OOM & OOM & OOM & OOM & OOM & OOM & OOM \\
        $\delta$-3-LWL & 5.163 & 1.930 & \bf 3.975 & \bf 2.277 & 1841.533 & 272.143 & 1411.924 & 126.022 & OOM & OOM & OOM & OOM \\
        $\delta$-3-LWL$^{+}$ & 5.151 & 1.931 & 60.375 & 41.110 & 1808.841 & 264.480 & 1346.608 & 123.394 & 380.480 & 19.073 & OOM & OOM \\
        \hline 
        \multicolumn{13}{c}{\textbf{Ridge, RBF}} \\
        \hline 
        SP & 58.721 & 34.606 & 60.375 & 41.110 & 131672.705 & 56058.593 & 181794.702 & 54604.564 & 38.945 & 14.712 & 5.474 & 2.224 \\
        GR & 11.670 & 5.663 & 12.488 & 5.012 & 42387.021 & 5110.985 & 41171.761 & 4831.495 & \bf 12.883 & \bf 5.073 & 4.804 & 1.944 \\
        WLOA & 58.719 & 34.635 & 25.906 & 17.002 & 92733.105 & 28242.033 & 137300.092 & 34067.513 & 32.827 & 12.230 & 3.215 & 1.261 \\
        WL & 58.719 & 34.635 & 25.905 & 17.003 & 109418.159 & 32350.523 & 112515.690 & 25035.268 & 26.313 & 10.933 & 2.227 & 0.837 \\
        2-WL & 11.010 & 5.926 & 12.618 & 8.317 & 40412.745 & 5351.789 & 21910.109 & 2982.532 & 32.424 & 12.948 & 7.164 & 3.271 \\
        $\delta$-2-WL & 10.500 & 4.630 & 9.316 & 6.207 & 40412.745 & 5351.789 & 21910.109 & 2982.532 & 32.518 & 13.045 & 7.409 & 3.287 \\
        $\delta$-2-LWL & 11.788 & 5.004 & 8.643 & 5.730 & 40412.745 & 5351.789 & 21910.109 & 2982.532 & 29.560 & 11.878 & 5.010 & 1.806 \\
        $\delta$-2-LWL$^{+}$ & 11.659 & 4.936 & 8.495 & 5.634 & 40412.745 & 5351.789 & 21910.109 & 2982.532 & 30.525 & 11.977 & 5.001 & 1.799 \\
        3-WL & 4.949 & 2.568 & 4.631 & 2.783 & OOM & OOM & OOM & OOM & 43.909 & 18.509 & OOM & OOM \\
        $\delta$-3-WL & 4.896 & 2.536 & 4.567 & 2.745 & OOM & OOM & OOM & OOM & 43.908 & 18.509 & OOM & OOM \\
        $\delta$-3-LWL & 16.720 & 2.980 & 5.356 & 3.149 & 89532.736 & 21918.757 & 91445.323 & 17703.656 & 43.909 & 18.510 & 10.925 & 5.320 \\
        $\delta$-3-LWL$^{+}$ & 16.721 & 2.972 & 60.375 & 41.110 & 89532.736 & 21918.757 & 91445.323 & 17703.656 & 43.908 & 18.513 & 12.488 & 6.501 \\
        \hline 
        \multicolumn{13}{c}{\textbf{Ridge, RBF w/ NIE}} \\
        \hline 
        SP & 58.721 & 34.606 & 60.375 & 41.110 & 131672.705 & 56058.593 & 181794.702 & 54604.564 & 38.945 & 14.712 & 5.474 & 2.224 \\
        GR & 11.670 & 5.663 & 12.488 & 5.012 & 42387.021 & 5110.985 & 41171.761 & 4831.495 & \bf 12.883 & \bf 5.073 & 4.804 & 1.944 \\
        WLOA & 58.719 & 34.635 & 25.906 & 17.002 & 31409.659 & 6644.798 & 19456.664 & 3892.678 & 24.429 & 10.354 & 3.163 & 1.189 \\
        WL & 58.719 & 34.635 & 25.905 & 17.003 & 48568.177 & 17533.158 & 71434.770 & 20472.124 & 23.155 & 9.302 & 2.026 & 0.805 \\
        2-WL & 11.010 & 5.926 & 12.618 & 8.317 & 28036.076 & 5266.623 & 48004.143 & 14046.171 & 34.729 & 14.580 & 8.301 & 3.679 \\
        $\delta$-2-WL & 10.500 & 4.630 & 9.316 & 6.207 & 15241.302 & 3289.949 & 48004.217 & 14047.425 & 34.707 & 14.584 & 8.266 & 3.669 \\
        $\delta$-2-LWL & 11.788 & 5.004 & 8.643 & 5.730 & 25849.115 & 4842.077 & 30846.779 & 6642.524 & 33.838 & 13.947 & 6.620 & 2.807 \\
        $\delta$-2-LWL$^{+}$ & 11.659 & 4.936 & 8.495 & 5.634 & 27368.926 & 5065.269 & 30093.401 & 6593.717 & 33.839 & 13.948 & 6.619 & 2.807 \\
        3-WL & 4.949 & 2.568 & 4.631 & 2.783 & OOM & OOM & OOM & OOM & OOM & OOM & OOM & OOM \\
        $\delta$-3-WL & 4.896 & 2.536 & 4.567 & 2.745 & OOM & OOM & OOM & OOM & OOM & OOM & OOM & OOM \\
        $\delta$-3-LWL & 16.720 & 2.980 & 5.356 & 3.149 & 856.975 & 160.003 & \bf 833.037 & \bf 75.286 & OOM & OOM & OOM & OOM \\
        $\delta$-3-LWL$^{+}$ & 16.721 & 2.972 & 60.375 & 41.110 & \bf 757.736 & \bf 148.417 & 886.330 & 88.512 & 43.918 & 18.491 & OOM & OOM \\
        \bottomrule
    \end{tabular}
\end{table*}

\end{document}